\documentclass[journal]{IEEEtran}

\usepackage{scalerel}
\usepackage{tikz}
\usetikzlibrary{svg.path}
\usepackage{xcolor}
\usepackage{amsmath}
\usepackage{amsfonts}
\usepackage{amssymb}
\usepackage{mathtools}
\usepackage[numbers]{natbib}
\usepackage[ruled,vlined,noend,linesnumbered]{algorithm2e}
\usepackage{caption}
\usepackage{subcaption}
\usepackage{multirow}
\usepackage{booktabs}
\usepackage{blkarray}
\usepackage[utf8]{inputenc}
\usepackage{fancyhdr}
\usepackage{pifont}
\newcommand{\cmark}{\ding{51}}%
\newcommand{\xmark}{\ding{55}}

\definecolor{orcidlogocol}{HTML}{A6CE39}
\tikzset{
    orcidlogo/.pic={
        \fill[orcidlogocol] svg{M256,128c0,70.7-57.3,128-128,128C57.3,256,0,198.7,0,128C0,57.3,57.3,0,128,0C198.7,0,256,57.3,256,128z};
        \fill[white] svg{M86.3,186.2H70.9V79.1h15.4v48.4V186.2z}
        svg{M108.9,79.1h41.6c39.6,0,57,28.3,57,53.6c0,27.5-21.5,53.6-56.8,53.6h-41.8V79.1z M124.3,172.4h24.5c34.9,0,42.9-26.5,42.9-39.7c0-21.5-13.7-39.7-43.7-39.7h-23.7V172.4z}
        svg{M88.7,56.8c0,5.5-4.5,10.1-10.1,10.1c-5.6,0-10.1-4.6-10.1-10.1c0-5.6,4.5-10.1,10.1-10.1C84.2,46.7,88.7,51.3,88.7,56.8z};
    }
}

\newcommand\orcidicon[1]{\href{https://orcid.org/#1}{\mbox{\scalerel*{
                \begin{tikzpicture}[yscale=-1,transform shape]
                \pic{orcidlogo};
                \end{tikzpicture}
            }{|}}}}

\usepackage{hyperref}

\def\@fnsymbol#1{\ensuremath{\ifcase#1\or *\or \dagger\or \ddagger\or
   \mathsection\or \mathparagraph\or \|\or **\or \dagger\dagger
   \or \ddagger\ddagger \else\@ctrerr\fi}}
\newcommand{\ssymbol}[1]{^{\@fnsymbol{#1}}}

\begin{document}
\title{EdgeTran: Co-designing Transformers for Efficient Inference on Mobile Edge Platforms}

\author{Shikhar~Tuli$^{\textsuperscript{\orcidicon{0000-0002-9230-5877}}}$,~\IEEEmembership{Student Member,~IEEE,} and~Niraj~K.~Jha,~\IEEEmembership{Fellow,~IEEE}
\thanks{This work was supported by NSF Grant No. CNS-2216746. S. Tuli and N. K. Jha are with the Department of Electrical and Computer Engineering,
Princeton University, Princeton, NJ, 08544, USA (e-mail: \{stuli, jha\}@princeton.edu).}
\thanks{Manuscript received ---; revised ---.}}

\markboth{}{Tuli \MakeLowercase{\textit{et al.}}: EdgeTran: Co-designing Transformers for Efficient Inference on Mobile Edge Platforms}


\maketitle

\begin{abstract}

Automated design of efficient transformer models has recently attracted significant attention from industry and 
academia. However, most works only focus on certain metrics while searching for the best-performing 
transformer architecture. Furthermore, running traditional, complex, and large transformer models on low-compute 
edge platforms is a challenging problem. In this work, we propose a framework, called ProTran, to profile the 
hardware performance measures for a design space of transformer architectures and a diverse set of edge devices. 
We use this profiler in conjunction with the proposed co-design technique to obtain the best-performing 
models that have high accuracy on the given task and minimize latency, energy consumption, and peak power draw 
to enable edge deployment. We refer to our framework for co-optimizing accuracy and hardware performance 
measures as EdgeTran. It searches for the best transformer model and edge device pair. Finally, we propose 
GPTran, a multi-stage block-level grow-and-prune post-processing step that further improves accuracy in a 
hardware-aware manner. The obtained transformer model is 2.8$\times$ smaller and has a 0.8\% higher GLUE score 
than the baseline (BERT-Base). Inference with it on the selected edge device enables 15.0\% lower latency, 
10.0$\times$ lower energy, and 10.8$\times$ lower peak power draw compared to an off-the-shelf GPU.

\end{abstract}


%
\IEEEpeerreviewmaketitle

\section{Introduction}

\IEEEPARstart{I}{n} recent years, self-attention-based transformer models~\citep{vaswani, bert} have achieved 
state-of-the-art results on tasks that span natural language processing (NLP) and, recently, even computer 
vision~\cite{vit_2021}. Increasing computational power and large-scale pre-training datasets have resulted in 
an explosion in transformer architecture size~\cite{turing_nlg}, much beyond state-of-the-art convolutional 
neural networks (CNNs). For instance, Megatron Turing-NLG~\cite{turing_nlg} has 530B trainable model parameters 
compared to only 928M trainable parameters in BiT (which uses ResNet-152 with every hidden layer widened by a 
factor of four, i.e., ResNet-152x4)~\cite{bit, resnet}. However, such massive transformer architectures are not 
amenable to operation on mobile edge devices due to a much lower compute budget and memory size. 

Another challenge of running large models on mobile devices is the extremely high latency 
incurred~\cite{hat_mit}. Smaller models may have reasonable latencies, however, they may still not meet the 
edge-level energy or peak power budgets. This could be due to a limited battery size or an intermittent power 
supply. Thus, there is a need for profiling and benchmarking latency, energy, and peak power consumption of
a diverse set of mobile-friendly transformer architectures. This would enable leveraging of hardware-aware 
neural architecture search (NAS)~\cite{hat_mit, flexibert} techniques to find the optimal architecture that 
maximizes model accuracy while meeting latency, energy, and peak power budgets.

Several works aim to prune transformer models to reduce the number of trainable model 
parameters~\cite{compressing_bert, micronet}. Some propose novel attention mechanisms to reduce the number 
of trainable parameters~\cite{fnet, linformer, squeezebert}. Others run NAS in a design space of transformer 
architectural hyperparameters to obtain efficient architectures~\cite{hat_mit, nas-bert, autotinybert}. However, 
most of these works only show gains in the number of model parameters or floating-point operations per second. Moreover, such works do not consider latency, energy, and power consumption in their 
architecture optimization loop. For instance, Wang et al.~\cite{hat_mit} only consider latency for running around 2000 transformer architectures on specific edge devices; Li et al.~\cite{ftrans} only target a single FPGA. 
Furthermore, the models obtained by such methods often have unacceptably high latencies, making 
them unusable for real-time NLP applications like machine translation. \textcolor{black}{Works that do consider hardware performance factors do not implement automated co-design with latency or energy consumption as feedback~\cite{wang2022towards}.} Thus, there is a need to profile not 
only the accuracy but also the latency, energy consumption, and peak power draw of transformer 
models on various mobile devices for inclusive design in edge-AI deployments under user-defined constraints.
However, profiling all models in a vast design space is a challenging endeavor. Hence, in this work, we make 
the following contributions.

\begin{itemize}
    \item We extend a previously proposed state-of-the-art benchmarking framework~\cite{flexibert}, called
FlexiBERT, to FlexiBERT 2.0. It uses an expanded design space of diverse transformer architectures for multiple 
edge-AI devices, targeting both training and inference. FlexiBERT 2.0 supports a finer-grained transfer of 
weights and increased heterogeneity compared to the original FlexiBERT framework, thus speeding up architecture 
search. It also supports a much more massive design space (10$^{79} \times $ larger) for mobile-friendly 
architectures, enabling a thorough search of the optimal architecture for the given edge platform. 
    \item We measure the latency, energy consumption, and peak power draw of the transformer models in our 
proposed design space. We call this profiling framework ProTran. It can obtain these hardware performance measures 
for a design space of transformer architectures on a given edge platform. It leverages an active-learning 
pipeline to \emph{efficiently} train surrogate models, minimizing the sample complexity of evaluation queries. 
It also supports various regression frameworks, including Gaussian process regression (GPR), decision tree 
(DT), boosted decision tree (BDT), and a state-of-the-art method, called BOSHNAS~\cite{flexibert} that exploits 
gradient-based optimization using backpropagation to inputs and heteroscedastic modeling~\cite{tuli2021cosco} 
to minimize overall uncertainty in the estimation of each measure. Using the proposed ProTran and FlexiBERT 
2.0 frameworks, any new edge device can be profiled within hours for subsequent quick transformer
architecture search under user-defined constraints.
    \item We then use ProTran's surrogate models and the proposed accuracy predictors as a foundation for
our \emph{fast} and \emph{efficient} co-design method for edge devices: EdgeTran. This co-design approach 
yields models with high accuracy but low latency, energy consumption, and peak power draw. Our co-design 
framework leverages \underline{B}ayesian \underline{o}ptimization using \underline{s}econd-order gradients and 
\underline{h}eteroscedastic surrogate modeling for \underline{co}-\underline{de}sign (BOSHCODE)~\cite{codebench}. It searches for 
optimal model-device pairs with the given constraints, \textcolor{black}{wherein it simultaneously chooses the edge device that performs best in terms of latency, energy consumption, and peak power draw while evaluating the searched transformer model architecture with high accuracy.} It can be used by academia and industry for scalable 
and streamlined deployments in a range of NLP tasks, targeting edge/cloud platforms.
    \item Finally, we propose a block-level multi-stage grow-and-prune post-processing step, GPTran, that further optimizes 
accuracy and hardware performance by adapting the structure of the converged transformer model.
\end{itemize}

We organize the rest of the article as follows. Section~\ref{sec:background} presents background on 
hardware-aware transformer design and deployment for efficient inference on edge platforms. 
Section~\ref{sec:methodology} illustrates the EdgeTran framework that includes FlexiBERT 2.0, ProTran, and GPTran. 
Section~\ref{sec:exp_setup} describes the experimental setup and the targeted baselines. 
Section~\ref{sec:results} presents the results. Section~\ref{sec:discussion} discusses the implications of the proposed work along with future work directions. Finally, Section~\ref{sec:conclusion} concludes the article.

\begin{figure*}[ht]
    \centering
    \includegraphics[width=\linewidth]{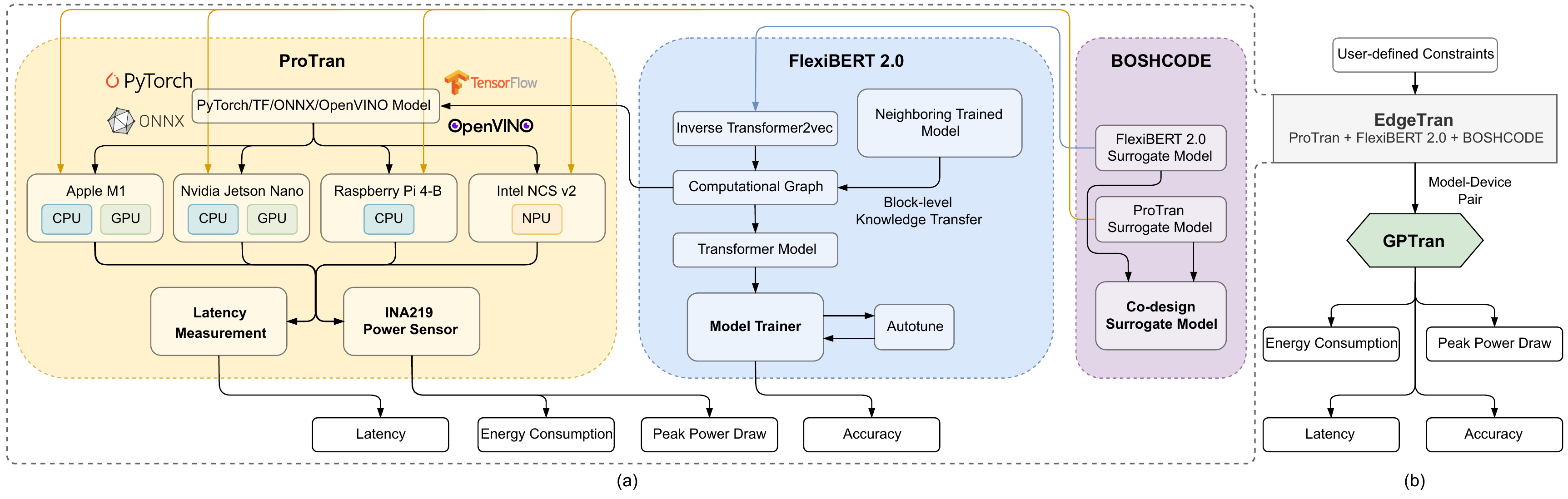}
    \caption{Overview of the EdgeTran framework: (a) ProTran used in conjunction with FlexiBERT 2.0 for modeling 
accuracy along with latency, energy consumption, and peak power draw (hardware measures) for different embedded 
platforms, using BOSHCODE~\cite{codebench} for co-design. (b) EdgeTran employs surrogate models obtained from ProTran and 
FlexiBERT 2.0 to obtain a best-performing model-device pair. We forward this model to GPTran for post-processing and 
further optimization.}
    \label{fig:edgetran_framework}
\end{figure*}

\section{Background and Related Work}
\label{sec:background}

This section introduces the relevant background and related works on hardware-aware NAS, pruning methods, and 
transformer architecture profiling.

\subsection{Transformer Architectures}

Previous works have proposed various transformer architectures. BERT is one of the most popular architectures 
that is widely used for language modeling~\cite{bert}. Its variants leverage mechanisms other than vanilla 
self-attention~\cite{shaw2018selfattention} to optimize performance or reduce model size and complexity. 
They include RoBERTa~\cite{roberta} that implements robust pre-training techniques, ConvBERT~\cite{convbert} 
that uses one-dimensional convolutional operations, MobileBERT~\cite{mobilebert} that uses bottleneck structures 
and multiple feed-forward stacks, SqueezeBERT~\cite{squeezebert} that uses grouped convolution operations to 
approximate the feed-forward stack, among others. Further, architectures like FNet~\cite{fnet} and 
LinFormer~\cite{linformer} use Fourier transform and low-rank approximation, respectively, of the self-attention operation to aid efficiency and reduce the number of model parameters. 

To obviate the need for hand-designed optimizations of the transformer model, many works devise design spaces 
to search for optimal architectural design decisions in a unified manner. For instance, SchuBERT~\cite{schubert} uses a design space of transformer architectures but does not consider different types of attention operations 
and only includes \emph{homogeneous} models (i.e., with the same encoder layer for every model) in the design space. DynaBERT~\cite{dynabert} adapts the width of the network by varying the number of 
attention heads and not the dimensionality of representation for each head. This only represents a simple 
extension to traditional architectures and does not target heterogeneity, much like other works that 
formulate design spaces for transformer architectures~\cite{nas-bert, autotinybert, adabert, autobertzero}. 

On the other hand, FlexiBERT~\cite{flexibert}, a state-of-the-art benchmarking framework for diverse transformer 
architectures, incorporates the most popularly used attention operations in a design space of 
\emph{heterogeneous} and \emph{flexible} transformer architectures. Each encoder layer in its design space can 
have a different attention mechanism (heterogeneity) and a different hidden dimension (flexibility). This leads to 
a vast design space consisting of 3.32 $\times$ 10$^9$ models, resulting in high gains in model performance 
for the same number of parameters. \textcolor{black}{The FlexiBERT surrogate model can also be used to predict the accuracy of any queried transformer in its design space (within reasonable uncertainty bounds). We provide more details on the validation performance of our surrogate model in Section~\ref{sec:exp_setup_surrogate_modeling}.} However, FlexiBERT's design space is not diverse enough to incorporate 
mobile-friendly architectures, has high training overhead while transferring weights to new models, and only 
supports the PyTorch platform, making it impractical for many edge devices. Nevertheless, taking inspiration 
from the advantages offered by flexible and heterogeneous models \textcolor{black}{and the benefits of expanding the search space to obtain better-performing models~\cite{codebench, pham2018efficient}}, we extend this framework to FlexiBERT 2.0
by targeting a more granular design space (more details in Section~\ref{sec:flexibert_design_space}). The 
FlexiBERT 2.0 framework enables us to model the latency, energy, and peak power draw of transformer 
architectures on a diverse set of embedded platforms. 

\subsection{Hardware-Aware Neural Architecture Search}

NAS techniques search for the architecture with the best accuracy in a specified dataset. However, NAS alone 
is hardly practical if we cannot run the best-performing transformer on the hardware at hand (or it does not meet hardware performance constraints). Recent state-of-the-art models, with billions of model parameters, 
exacerbate this problem~\cite{turing_nlg}. Hence, recent works have focused on hardware-aware NAS,  
directed at architecture search for a target platform. For example, ChamNet proposed accuracy and resource 
(latency and energy) predictors and leveraged GPR-based Bayesian optimization (GPR-BO) to find the optimal 
CNN architecture for a given platform~\cite{chamnet}. Some works have proposed \emph{co-design} of hardware and 
software design decisions~\cite{codebench, naas, mcunet}. However, they are limited to CNN design spaces.

HAT~\cite{hat_mit}, a recent framework for hardware-aware NAS for transformers, trains a large transformer model 
first and then uses latency feedback to obtain a sub-model for the given hardware platform. However, all 
sub-models are homogeneous (in terms of attention type) and have constant dimensionality in each encoder layer, 
limiting their representation capacity~\cite{flexibert}. Further, this work uses a static training recipe, which 
may not be optimal for every sub-model. Lastly, as recent works have shown~\cite{flexibert}, its design space is highly restricted. Instead, one can leverage other NAS techniques for superior and efficient 
search of the optimal model in a diverse set of transformer architectures~\cite{flexibert, chamnet, nest}. 

Fig.~\ref{fig:edgetran_framework}(a) shows how ProTran leverages the proposed FlexiBERT 2.0 framework to obtain 
various hardware performance measures for diverse architectures. First, we convert each queried model in the 
FlexiBERT 2.0 design space to a computational graph that we train (while also autotuning the training recipe 
to improve accuracy further). \textcolor{black}{FlexiBERT 2.0 supports a range of deep learning frameworks,
including PyTorch~\cite{pytorch}, TensorFlow~\cite{tensorflow}, ONNX~\cite{onnx}, and OpenVINO~\cite{openvino}
(see Section~\ref{sec:model_formats}). Thus, one can profile any new hardware supported by any of these frameworks.} 
We then pass this model to the ProTran framework that runs inference for
different NLP tasks on diverse mobile platforms. These platforms include Apple M1 with both a central processing 
unit (CPU) and an embedded graphics processing unit (GPU), Raspberry Pi embedded CPU, Intel Neural Compute 
Stick (NCS) v2 that has an embedded neural processing unit (NPU), and the Nvidia Jetson Nano (that has a CPU and an
embedded GPU). We provide more details on the selected set of mobile platforms along with server-side baselines 
in Section~\ref{sec:exp_setup_hw_platforms}. ProTran then outputs the latency, energy consumption, and peak 
power draw of the given transformer model that EdgeTran can use for hardware-aware NAS and co-design. Next, 
BOSHCODE queries the FlexiBERT 2.0 and ProTran frameworks to create surrogate models for model accuracy and 
hardware performance (latency, energy consumption, and peak power draw) for the selected set of hardware 
platforms. \textcolor{black}{We train these surrogate models as a pre-processing step so that one does not have to train or run inference on the target hardware multiple times. This enables faster search using lightweight surrogate models.} 

Fig.~\ref{fig:edgetran_framework}(b) presents EdgeTran, which exploits 
surrogate models obtained from ProTran and FlexiBERT 2.0. It runs co-design using the BOSHCODE framework to output 
an optimal model-device pair for the set of user-defined constraints. Finally, we input this pair to the 
GPTran post-processing step to optimize the transformer model further and improve performance.

\subsection{\textcolor{black}{Latency and Energy Profiling of Transformer Models}}

\textcolor{black}{Model latency is evaluated when running a batch of input with a given transformer model architecture. Wang et al.~\cite{wang2022towards} measure the latency of transformer inference on the Pixel 4 smartphone. However, the inference latency on certain edge devices can go up to hundreds of seconds. Thus, there is a need for a lightweight surrogate model that can quickly predict model inference latency (in a few milliseconds). We train this surrogate model for latency, energy, and peak power estimation of diverse models in the FlexiBERT 2.0 design space using the proposed ProTran framework.}

Energy consumption profiling of a machine learning (ML) 
model is challenging. This is because extracting the energy consumed only by training or running inference 
processes for an ML model is nontrivial. Further, when the design space is enormous, running training or 
inference for each model may incur drastically long runtimes. Nevertheless, previous works have profiled the 
energy consumption of ML architectures. For example, ChamNet~\cite{chamnet} trains energy predictors 
for various CNNs in its design space on different hardware platforms under various latency constraints. 
FTRANS~\cite{ftrans} obtains energy and power consumption for different transformer architectures on an FPGA. 
Some works have attempted to co-optimize hardware and transformer, however, under a very limited 
scope~\cite{accelerating_framework, spatten}. These works prune the weights of a given model to reduce model 
complexity. However, the total model size remains significant. This calls for a rigorous search of inherently 
dense but smaller architectures that one could run on the device with a minimal memory footprint. This search 
falls under the domain of hardware-aware NAS. However, to the best of our knowledge, no transformer NAS 
approach has simultaneously accounted for accuracy, latency, energy consumption, and peak power 
draw~\cite{hat_mit, flexibert, ftrans, wang2022towards, chamnet, accelerating_framework}. Thus, there is a need for lightweight surrogate models 
for the estimation of these measures on a diverse set of transformer architectures for various edge-AI 
platforms. This would enable energy-aware NAS of transformer models and efficient co-design for optimal edge 
deployments. Finally, we could extend it to search for optimal transformer-accelerator pairs~\cite{naas, energon}.

\section{Methodology}
\label{sec:methodology}

In this section, we first present FlexiBERT 2.0 extensions relative to its predecessor. We also describe the 
ProTran pipeline for measuring hardware performance on diverse edge-AI platforms. We then present the 
BOSHCODE co-design method. Finally, we give details on the proposed GPTran framework for optimizing transformer 
architectures.

\subsection{FlexiBERT 2.0 Framework}
\label{sec:flexibert_2}

We discuss the FlexiBERT 2.0 design space next.

\subsubsection{Design Space}
\label{sec:flexibert_design_space}

\begin{table}[t]
\caption{Design space description. Super-script ($j$) depicts the value for layer $j$.}
\vskip 0.1in
    \centering
    \resizebox{\columnwidth}{!}{
    \begin{tabular}{@{}ll@{}}
    \toprule
        \textbf{Design Element} & \textbf{Allowed Values} \\
        \midrule 
        Number of encoder layers ($l$) & $\{2,4,6,8,10,12\}$\\
        Type of attention operation used ($o^j$) & $\{\text{SA},\text{LT},\text{DSC}\}$\\
        Number of operation heads ($n^j$) & $\{2,4,8,12\}$ \\
        Hidden size ($h^j$) & $\{128,256,512,768\}$\\
        Feed-forward dimension ($f^j$) & $\{256,512,1024,2048,3072,4096\}$\\
        Number of feed-forward stacks & $\{1,2,3\}$\\
        Operation parameters ($p^j$): & \\
       \hspace{3mm} if $o^j = \text{SA}$ & Self-attention type: $\{\text{SDP},\text{WMA}\}$ \\ 
       \hspace{3mm} else if $o^j = \text{LT}$ & Linear transform type: $\{\text{DFT}, \text{DCT}\}$ \\
       \hspace{3mm} else if $o^j = \text{DSC}$ & Convolution kernel size: $\{5,9\}$ \tabularnewline
       \bottomrule
    \end{tabular}}
    \label{tab:hyp_ranges}
\end{table}

The traditional BERT model~\cite{bert} consists of multiple layers, 
each containing a bidirectional multi-headed self-attention (SA) module followed by a feed-forward module 
(that implements a fully-connected neural network with a single hidden layer). Searching over a space of BERT-like homogeneous models results in marginal gains~\cite{schubert}. However, as proposed by Tuli et al.~\cite{flexibert}, the design space of \emph{heterogeneous} transformer architectures is immense because one can add a diverse set of operations to them. \textcolor{black}{Such scale and diversity enable a rigorous search for the best-performing model resulting in significant gains over traditional search spaces~\cite{flexibert, codebench}}. Due to these advantages, we leverage the expansive, heterogeneous, yet modular FlexiBERT 
architectures in our design space. We propose several modifications 
to the original BERT encoder layer, primarily to the attention module. 

\begin{figure}
    \centering
    \includegraphics[width=0.8\linewidth]{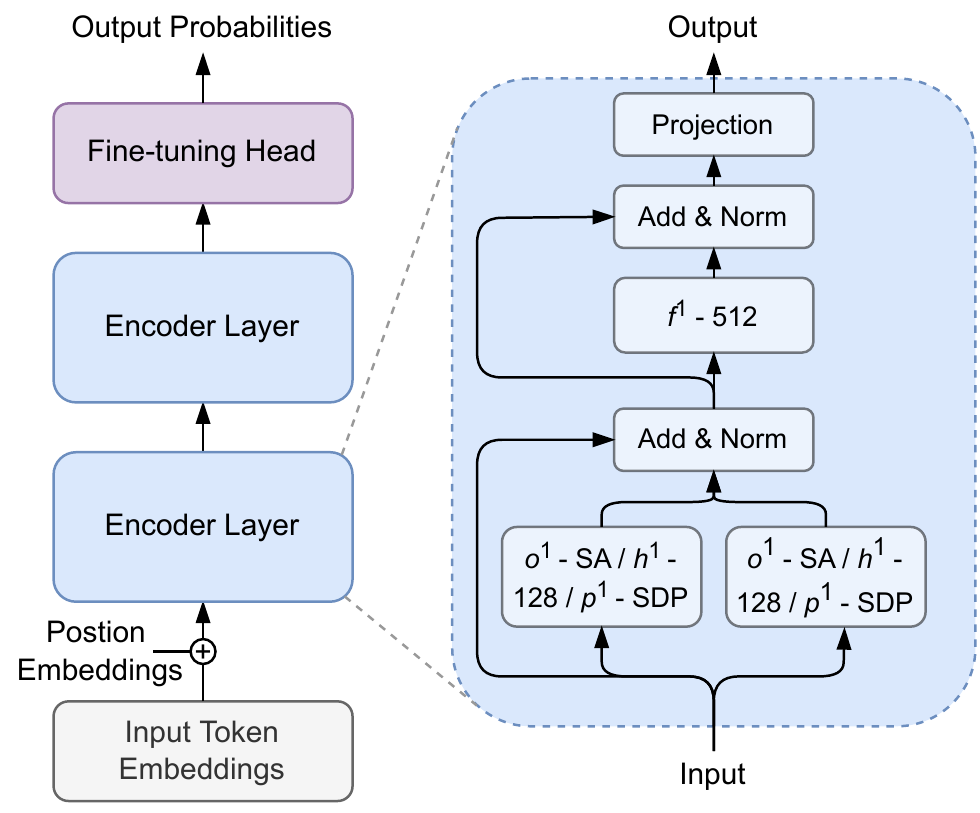}
    \caption{BERT-Tiny in the FlexiBERT 2.0 representation.}
    \label{fig:bert_tiny_comp_graph}
\end{figure}

We consider weighted multiplicative attention (WMA)-based self-attention~\cite{wma} in addition to scaled 
dot-product (SDP)-based self-attention. We also incorporate linear transform (LT)-based attention in 
FNet~\cite{fnet} and dynamic-span-based convolution (DSC) in ConvBERT~\cite{convbert}, in place of the vanilla 
self-attention mechanism. Whereas the original FNet implementation uses discrete Fourier transform (DFT), we 
also consider discrete cosine transform (DCT) in our design space. We further allow variable kernel sizes for 
convolution-based attention. Consolidating different attention module types (also called operations) that vary in their computational 
costs into a single design space enables the models to have inter-layer variance in expression capacity. 
Inspired by MobileBERT~\cite{mobilebert}, we consider architectures with multiple fully-connected layers (we 
call this a feed-forward stack). We summarize the entire design space with the range for each operation type 
in Table~\ref{tab:hyp_ranges}. Considering all the possible hyperparameter values given in 
Table~\ref{tab:hyp_ranges}, we generate a design space with $1.69 \times 10^{88}$ architectures, much larger 
than those in any previous work. Fig.~\ref{fig:bert_tiny_comp_graph} shows how we represent the BERT-Tiny 
model~\cite{turc2019} in our proposed design space.

Unlike the original FlexiBERT design space, FlexiBERT 2.0 uses a broader range of values for each hyperparameter 
to target even more diverse architectures. This results in models that are substantially different from 
traditional BERT-like ones~\cite{bert, fnet, convbert}. Further, we make the architectures in the FlexiBERT 2.0 
pipeline even more \emph{heterogeneous}, i.e., instead of all attention operations in an encoder layer being 
the same~\cite{flexibert}, it allows an encoder layer to have different types of operations. For instance, where 
the original FlexiBERT only allows SA heads in an encoder layer, our design space also allows other 
attention types (from WMA, DCT, DFT, DSC) in the same layer. Each attention head is also allowed a different 
hidden dimension. 

\subsubsection{Weight Transfer}
\label{sec:weight_transfer}

\textcolor{black}{To obtain the accuracy of a new model (queried by the active learning framework to train the surrogate model), training it from scratch would be computationally expensive. FlexiBERT~\cite{flexibert} implements weight transfer at the `encoder-level' so that queried models are not trained from scratch. It transfers weights from a neighboring pre-trained model. This speeds up the training of queried models. In the proposed framework, we not only leverage weight transfer to train the surrogate model quickly but also for training new models while implementing the GPTran pipeline (details in Section~\ref{sec:gptran_framework}).}

\begin{figure}
    \centering
    \includegraphics[width=0.9\linewidth]{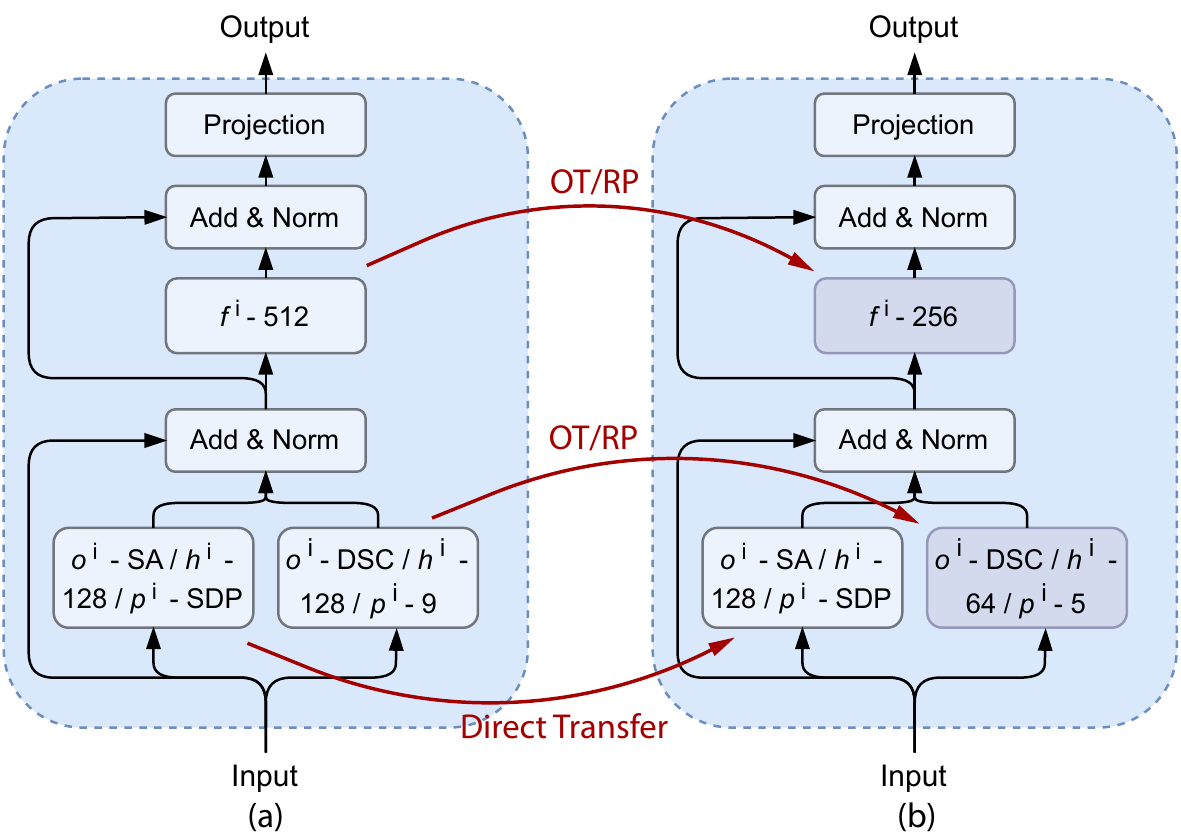}
    \caption{Weight transfer between two neighboring models in FlexiBERT 2.0.}
    \label{fig:weight_transfer}
\end{figure}

We update the original FlexiBERT's weight transfer to an `attention-head level,' i.e., we do not require the entire encoder 
layer hyperparameters to be the same for transferring the weights. If some of the attention heads are alike in 
two neighboring models, the weights for those attention heads can be \emph{directly} transferred. When attention heads have different dimensions but the rest of the parameters are the 
same, we implement weight transfer by cloning an ordered part of the weight matrix [we call this ordered 
transfer (OT)] or by random projection (RP) of the original weight matrix to the new dimension. RP takes 
inspiration from dimensionality reduction techniques based on the Johnson-Lindenstrauss lemma~\cite{rp}. To 
implement RP, we project the original input space on a randomly generated matrix. We draw its components from 
a Gaussian distribution $\mathcal{N}(0, 1/n_c)$, where $n_c$ is the number of components or the dimensionality of 
the target space. This method decreases the loss of information when transferring weights to a lower dimension, 
reducing the number of iterations to train the new neighbor. This, in turn, lowers the net training time of all 
queries, reducing the overall search time. Fig.~\ref{fig:weight_transfer} summarizes the weight transfer process 
in FlexiBERT 2.0 for two neighboring models.

\subsubsection{Support for Model Formats}
\label{sec:model_formats}

To enable profiling on diverse edge-AI platforms, FlexiBERT 2.0 supports various ML frameworks. All models in 
the FlexiBERT 2.0 design space can be saved in PyTorch, Tensorflow, 
ONNX, and OpenVINO formats. This broadens the scope of our proposed models to a 
wide range of platforms, enabling unified benchmarking and wider deployments.

\subsubsection{Transformer Embeddings}
\label{sec:embeddings}

The original FlexiBERT pipeline leverages a \texttt{Transformer2vec} embedding scheme to form dense embeddings 
for the transformer architectures in the design space. However, training these embeddings is computationally 
expensive. We thus propose an embedding scheme that does not require training. We illustrate this scheme next.

For the selected ranges of hyperparameters in our design space (see Table~\ref{tab:hyp_ranges}), we generate 
37-dimensional embeddings as follows:

\begin{itemize}
    \item The first entry in the embedding is the number of encoder layers ($l$) in the current transformer 
model. In other words, for the embedding of a transformer architecture $e$, $e_1$ represents the number of 
encoder layers in the architecture.
    \item For an encoder layer $j$, $e_{2 + 3(j-1)}$ represents the hidden dimension ($h^j$). This is the sum 
of the hidden dimension for each attention head in that encoder layer. Other embedding indices determine how 
much $h^j$ we allocate to each attention head.
    \item For an encoder layer $j$, $e_{3 + 3(j-1)}$ represents the index of the feed-forward operation formed 
by the range of feed-forward layers possible in the given design space. For the six possible hidden dimensions 
in the feed-forward layers (see Table~\ref{tab:hyp_ranges}), there can be a stack of up to three layers, thus 
giving rise to 258 feed-forward operation types for every encoder layer.
    \item For an encoder layer $j$, $e_{4 + 3(j-1)}$ represents the index of the attention head operation in 
the list of multi-head operations types. We obtain this list based on the number of attention heads selected 
for that encoder layer, the type of each attention head, the hidden dimension for each attention head, and 
their respective combinations with replacement (more details in Section~\ref{sec:exp_setup_flexibert_2}).
    \item For models less than 12 layers deep, we set the respective entries in their embeddings to zero.
\end{itemize}

Although these embeddings are sparse, they are much faster to obtain than training with the 
\texttt{Transformer2vec} embeddings~\cite{flexibert}. This is especially important due to the much larger design 
space of the proposed framework.

\subsection{The ProTran Framework}

We now describe the ProTran framework that leverages the FlexiBERT 2.0 design space to train surrogate models 
for latency, energy consumption, and peak power draw on diverse platforms. We train the surrogate models in 
an active-learning fashion~\cite{al_survey}. We first obtain an initial set of \emph{diverse} samples to 
initialize our surrogate model. Then, we use the \emph{uncertainty} estimates from that model to query new 
architectures until the maximum uncertainty falls below a threshold.

\subsubsection{Initial Sampling}

\begin{figure}
    \centering
    \includegraphics[width=\linewidth]{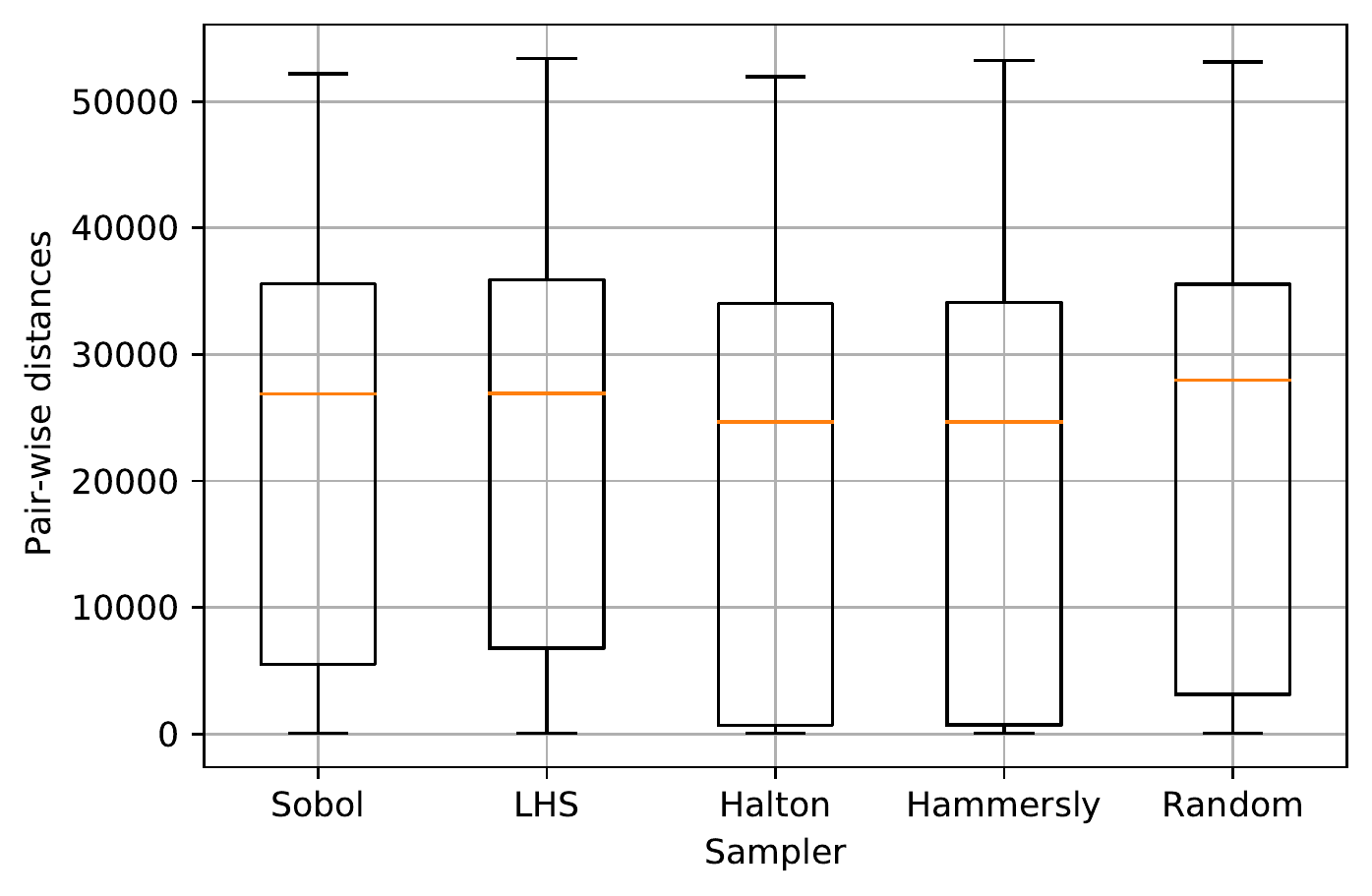}
    \caption{Box plot for pairwise distances of 256 sampled embeddings from different sampling schemes.}
    \label{fig:sampler_boxplot}
\end{figure}

Any regressor used for an active-learning pipeline requires an initial seed dataset to predict further queries it needs to explore. Intuitively, this dataset should be as representative of the design space as possible. 
For this, we test various low-discrepancy sequence sampling strategies~\cite{random_number_generation}. 
Fig.~\ref{fig:sampler_boxplot} shows a boxplot of pairwise distances between embeddings using various sampling 
methods, namely, Sobol sampling, Latin hypercube sampling (LHS), Halton sampling, Hammersly sampling, and 
Random sampling. We use the LHS method in our experiments since it results in the highest first quartile of 
pairwise distances between the embeddings of the sampled architectures. In other words, it maximizes the probability of 
having divergent points in the sampled set.

\begin{figure}
    \centering
    \includegraphics[width=0.9\linewidth]{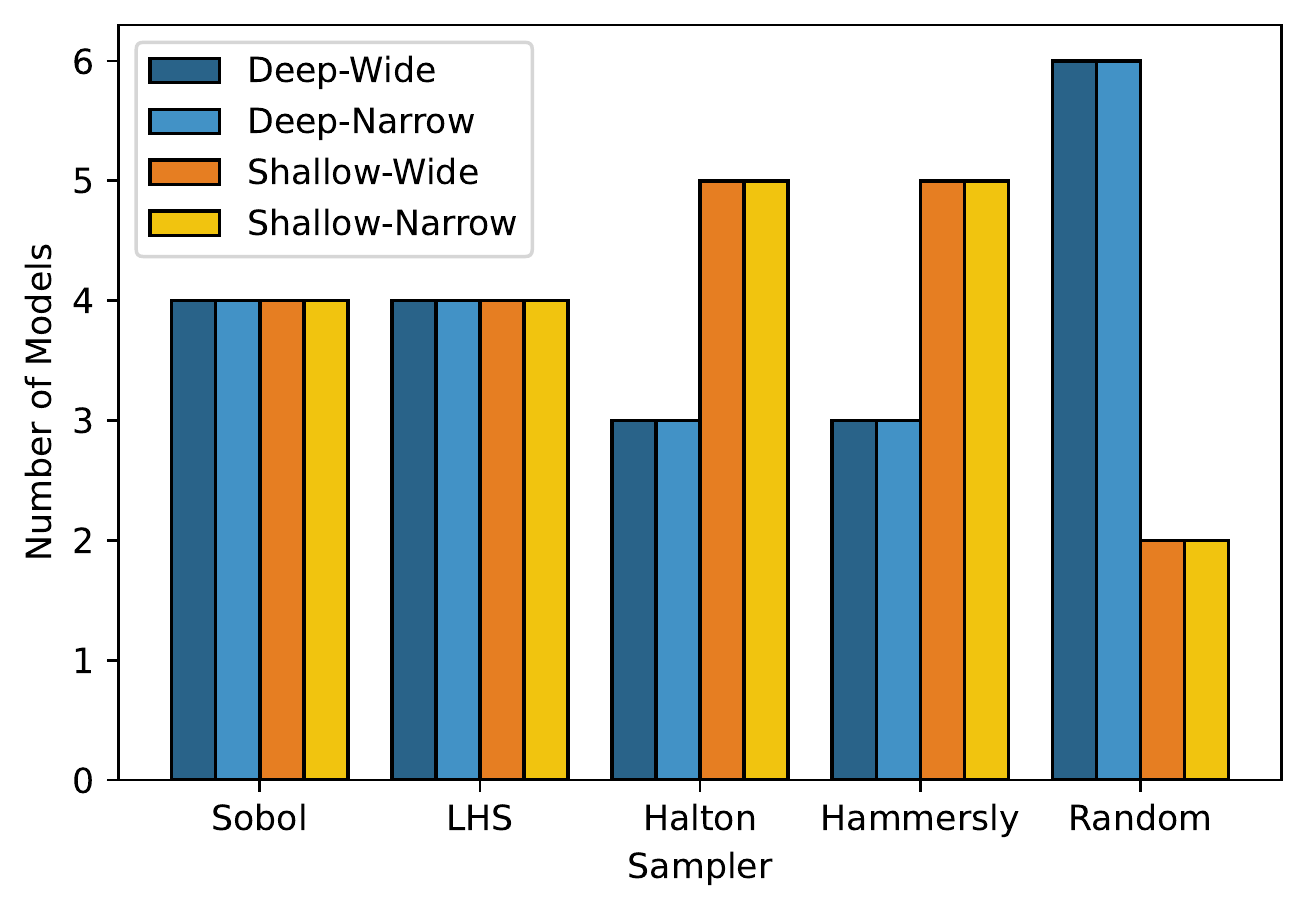}
    \caption{Model diversity using various sampling schemes.}
    \label{fig:diversity_barplot}
\end{figure}

We obtain 16 samples using the chosen method to initialize the seed dataset. To further test for sample diversity, 
we segregate the sampled models into four categories: deep-wide, deep-narrow, shallow-wide, and 
shallow-narrow. The model is shallow if the number of encoder layers is strictly less than eight and deep otherwise. The 
model is narrow if the median number of attention heads is strictly less than eight and wide otherwise. 
Fig.~\ref{fig:diversity_barplot} shows the number of each model type obtained using each sampling scheme with 
16 initial samples. Sobol and LHS methods have equidistribution of model types, demonstrating high 
diversity~\cite{diversity_measures} relative to other schemes. 

\subsubsection{Learning the Performance Predictor}
\label{sec:learning_perf_pred}

\begin{figure}
    \centering
    \includegraphics[width=0.9\linewidth]{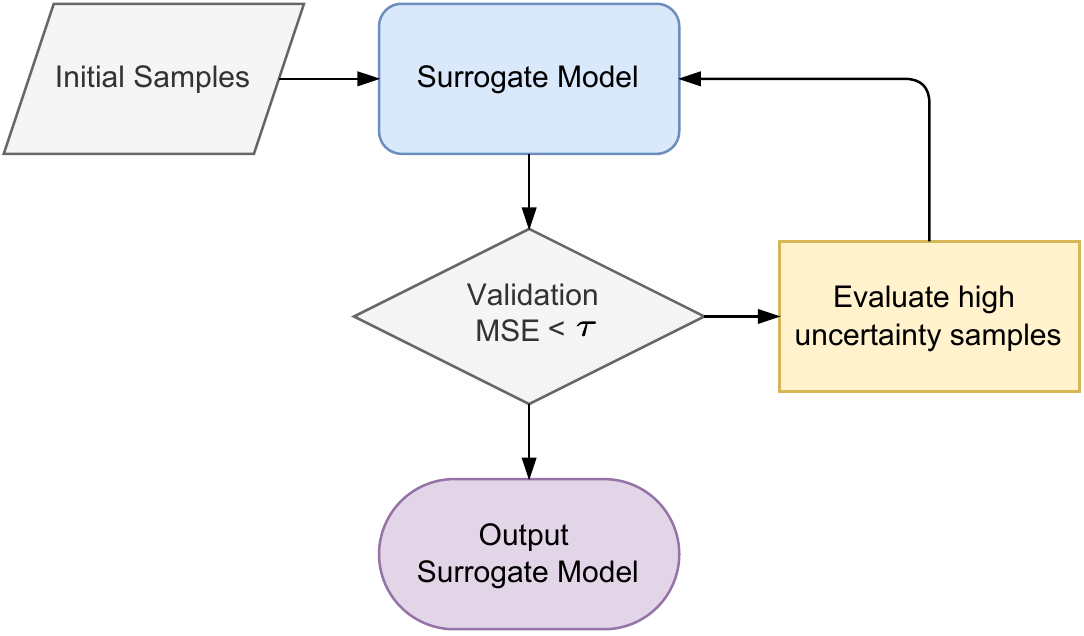}
    \caption{The active-learning pipeline of ProTran.}
    \label{fig:protran_pipeline}
\end{figure}

Once we have evaluated the initial samples for all the hardware performance measures on a given edge-AI
platform (see Section~\ref{sec:exp_setup_power_meas}), we need to learn a performance predictor that takes
the transformer embedding as input and predicts each measure under an error constraint. We can
eventually leverage this predictor, also called a surrogate model, along
with a corresponding uncertainty estimation, to query novel models in the design space. This further increases
confidence in estimation (i.e., lowers the uncertainty or validation error). We employ this strategy in an
active-learning fashion to minimize the number of queried models for evaluation. Here, by validation error, we mean 
the error of the predictor on untrained samples. Fig.~\ref{fig:protran_pipeline} shows a flowchart of this 
pipeline. We use the initial 16 LHS-sampled transformer architectures to initialize the surrogate model. We then 
evaluate high-uncertainty samples to train the surrogate model iteratively on the expanding dataset until the 
validation mean-squared error (MSE) falls below a predetermined threshold.

\begin{figure}
    \centering
    \includegraphics[width=\linewidth]{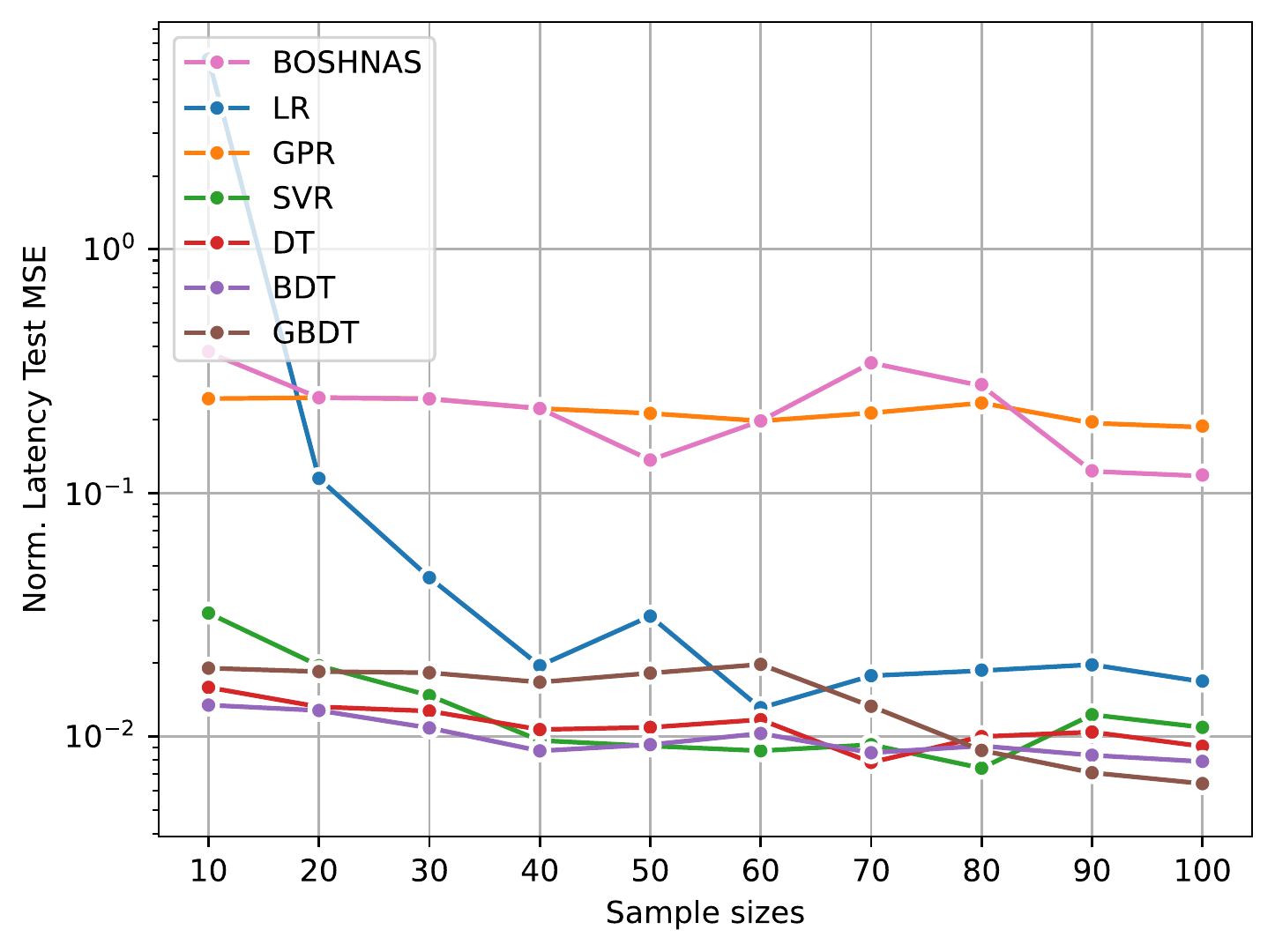}
    \caption{Validation MSE on the normalized latency values for different sample sizes while using various 
regressors for the A100 GPU.}
    \label{fig:latency_test_mse}
\end{figure}

\begin{table}[t]
\caption{Selected batch size and the number of samples for convergence for different platforms.}
\centering
\begin{tabular}{@{}l@{\hskip .3in}|@{\hskip .3in}c@{\hskip .3in}c@{}}
\toprule
Platform & Batch size & Number of samples \\ \midrule
Nvidia A100 GPU & 128 & 239 \\ [1mm]
Apple M1 CPU & 32 & 104 \\ [1mm]
Apple M1 GPU & 32 & 92 \\ [1mm]
Raspberry Pi CPU & 1 & 81 \\ [1mm]
Intel NCS NPU & 1 & 21 \\ [1mm]
Nvidia Jetson Nano CPU & 1 & 223 \\ [1mm]
Nvidia Jetson Nano GPU & 1 & 22 \\ \bottomrule
\end{tabular}
\label{tab:sample_sizes}
\end{table}

For the active-learning loop, we experiment with several regression schemes, i.e., surrogate models,
namely linear regression (LR), GPR, support-vector regression (SVR), DT, BDT, gradient-boosted decision trees (GBDT), 
and BOSHNAS~\cite{flexibert} that exploits gradient-based optimization using
backpropagation to the input and heteroscedastic modeling~\cite{tuli2021cosco}. We employ these models to
minimize the overall uncertainty in estimating the prediction measures. GPR and BOSHNAS directly indicate
the epistemic uncertainty in predictions; thus, they select the following query in the active-learning loop as 
the model with the highest uncertainty. We compute the uncertainty in estimation for BDT and GBDT as the 
standard deviation in the predictions of each decision tree for every output hardware performance measure. 
However, LR, DT, and SVR cannot model uncertainty in performance prediction. In such cases, we evaluate 
random samples to expand the dataset.

We test these regressors to model the inference latency on the Nvidia A100 GPU 
for the SST-2 task in the GLUE benchmarking suite~\cite{glue}. For a pool of high-uncertainty
samples evaluated at each iteration, we take smaller subsets of train/validation (80-20\%) splits to check the 
prediction MSE on the validation set after training the regressor on the training set. From
Fig.~\ref{fig:latency_test_mse}, we see that GBDT reaches the lowest prediction error on the validation set as we increase
the sample size. BOSHNAS does not perform well due to the high sample sizes required to train neural
network surrogates optimally~\cite{nasbench_301}. Thus, we choose GBDT as our surrogate model in the
active-learning loop while training performance predictors for all platforms. Table~\ref{tab:sample_sizes} 
shows the sample sizes required for GBDT to converge for different platforms. We reach convergence when 
the validation MSE falls below 0.5\% for latency, energy, and peak power draw, individually, when normalized 
(e.g., we divide the latency values by the maximum latency encountered in the dataset in order to obtain 
normalized values between 0 and 1).

\begin{figure}[t]
    \centering
    \includegraphics[width=0.7\linewidth]{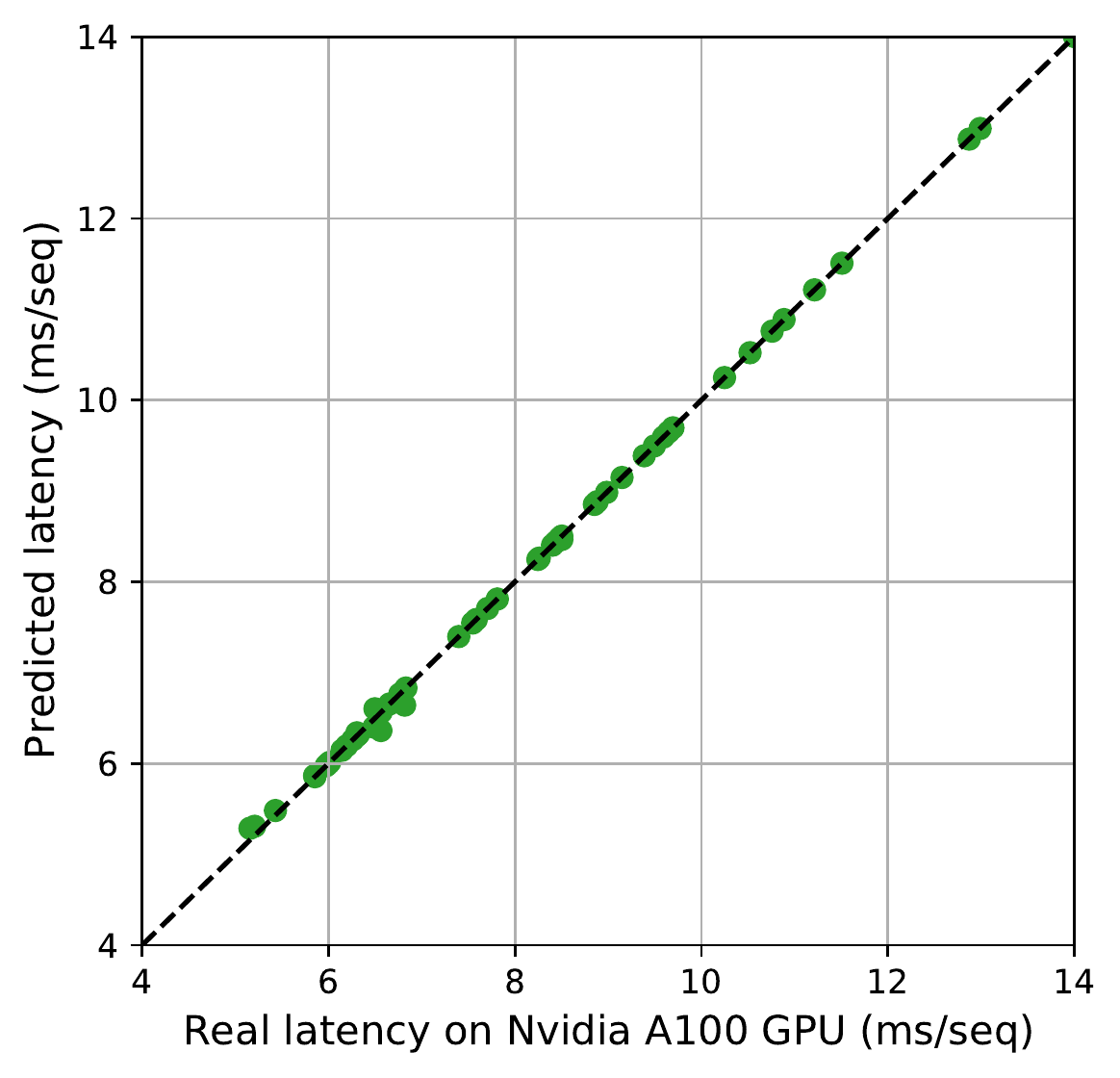}
    \caption{Predicted and real latencies on the Nvidia A100 GPU. Latencies are reported per sequence in the 
SST-2 task for transformer model evaluation.}
    \label{fig:pred_vs_real_latency_bdt_a100}
\end{figure}

Fig.~\ref{fig:pred_vs_real_latency_bdt_a100} shows the predicted latency of 32 sampled transformer models
obtained using the GBDT regressor against the real latency on the Nvidia A100 GPU. The plot shows that the 
predicted latency is very close to the real latency.

\subsection{BOSHCODE}

BOSHNAS~\cite{flexibert} is a NAS technique that runs gradient-based optimization using backpropagation to the 
input (GOBI)~\cite{tuli2021cosco} on a \emph{single} and \emph{lightweight} neural network (NN) model that 
predicts not only model performance, but also the epistemic and aleatoric uncertainties. It leverages an 
active-learning framework to optimize the upper confidence bound (UCB) estimate of model performance in the 
embedding space. Estimates of aleatoric uncertainty enable further optimization of the training recipe for every 
model in the design space. GOBI freezes the model weights and backpropagates the gradients towards 
the input values to minimize the output optimization measure~\cite{tuli2021cosco}. We extend the application of 
BOSHNAS to BOSHCODE~\cite{codebench}, a co-design framework for transformer models and edge-AI devices. We describe
this framework next.

\subsubsection{Uncertainty Types}

Prediction uncertainty can arise from not only the approximations in the surrogate 
modeling process but also parameter initializations and variations in model performance due to different 
training recipes. They are referred to as \emph{epistemic} and \emph{aleatoric} uncertainty, respectively. 

\subsubsection{Surrogate Model}

Following the surrogate modeling approach used in CODEBench~\cite{codebench}, a co-design method for CNNs and accelerators, 
we model the performance and the aleatoric uncertainty using a natural parameter network (NPN)~\cite{npn} 
$f(x_\text{TXF}, x_\text{ED}; \theta)$. We model the epistemic uncertainty using $g(x_\text{TXF}, x_\text{ED}; \theta')$ 
and $h(x_\text{TXF}, x_\text{ED}; \theta'')$. We leverage GOBI on $h$, a student network for the teacher $g$, to avoid numerical gradients due to their poor performance~\cite{codebench}. Here, $x_\text{TXF}$ refers to the transformer embedding and $x_\text{ED}$ 
refers to the embedding for the edge device ($\theta$, $\theta'$, and $\theta''$ refer to the training parameters of the 
models). $(\mu, \sigma) \gets f(x_\text{TXF}, x_\text{ED}; \theta)$, where $\mu$ is the predicted mean performance and $\sigma$ is the aleatoric uncertainty. Moreover, $h$ predicts a surrogate ($\hat{\xi}$) of the epistemic uncertainty ($\xi$)~\cite{codebench}.

\begin{figure}
    \centering
    \includegraphics[width=\linewidth]{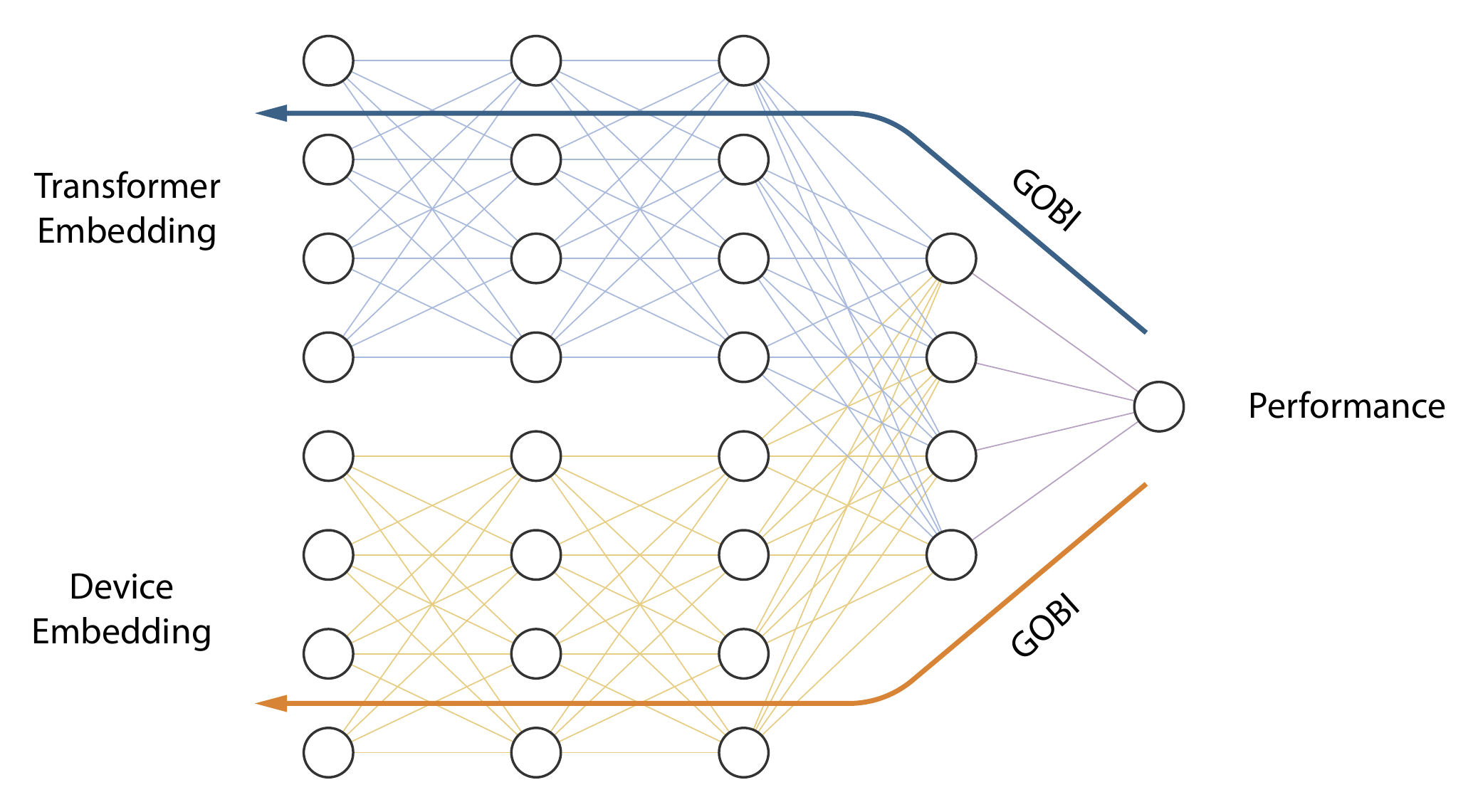}
    \caption{Teacher network in the BOSHCODE surrogate model. Dropout layers have been omitted for simplicity~\cite{codebench}.}
    \label{fig:boshcode_teacher}
\end{figure}

Fig.~\ref{fig:boshcode_teacher} shows a simplified schematic of the teacher network $g$ in BOSHCODE~\cite{codebench}. It realizes the
model-device embeddings, a combination of the 37-dimensional transformer embeddings (see Section~\ref{sec:embeddings}) and 7-dimensional one-hot device encodings. 
We run GOBI on the combined and separate representations (of the student network $h$~\cite{codebench}) to find the optimal model-device 
pair that maximizes the UCB estimate of the performance ($P$). Here, performance refers to a convex combination of 
model accuracy and hardware performance measures (latency, energy, and peak power consumption). Mathematically, 
\begin{align}
\label{eqn:perf_metric}
\begin{split}
    &\text{Performance } (P) = \alpha \times \text{Accuracy} \\
    &+ \beta \times (1 - \text{Energy Consumption}) \\
    &+ \gamma \times (1 - \text{Peak Power Draw}) + \epsilon \times (1 - \text{Latency})
\end{split}
\end{align}
where $\alpha + \beta + \gamma + \epsilon = 1$ are hyperparameters. We normalize the values of the individual performance 
measures with respect to their maximum values (thus these values reside in the $[0,1]$ interval). 
For different applications, the user can define constraints based on the values of these hyperparameters. 
For instance, if accuracy is of utmost importance, $\alpha$ can be set high. On the other hand, in real-time machine translation 
applications that require low latency, $\epsilon$ can be set high.

\subsubsection{Active Learning and Optimization}

In a design space of model-device pairs $\Delta$, we search for the predicted best-performing pairs in an 
active-learning fashion. Assuming we have the three networks $f, g$, and $h$ initialized based on a randomly 
sampled set of model-device pairs ($\delta$), we run second-order optimization on $\mathrm{UCB} = \mu + k_1 \cdot \sigma + k_2 \cdot \hat{\xi}$~\cite{codebench},
where $x_\text{TXF}, x_\text{ED} \in \Delta$, $k_1$, and $k_2$ are hyperparameters. 

\SetKwComment{Comment}{/* }{ */}
\begin{algorithm}[t]
\SetAlgoLined
\KwResult{trained \textbf{surrogate} model}
 \textbf{Initialize:} convergence criterion, uncertainty sampling 
prob. ($\alpha_P$), diversity sampling prob. ($\beta_P$), \textbf{surrogate} model ($f$, $g$, and 
$h$) on initial sample set $\delta$, design space $[x_\text{TXF}, x_\text{ED}] \in \Delta$; \\
 \While{convergence criterion not met}{
  \eIf{prob $\sim U(0,1) < 1 - \alpha_P - \beta_P$}{
  fit(\textbf{surrogate}, $\delta$)\; \label{line:fit}
    $x_\text{TXF}, x_\text{ED}$ $\gets$ GOBI($f$, $h$) \Comment*{Optim.} \label{line:opt}
    EVALUATE($x_\text{TXF}, x_\text{ED}$)\; \label{line:eval}
  }{
  \eIf{$1 - \alpha_P - \beta_P \le$ prob. $< 1 - \beta_P$}{
    $x_\text{TXF}, x_\text{ED}$ $\gets$ $\underset{x_\text{TXF}, x_\text{ED}}{\textbf{argmax}}$($k_1 \cdot \sigma + k_2 \cdot \hat{\xi}$)  \Comment*{Uncertainty sampling}
    EVALUATE($x_\text{TXF}, x_\text{ED}$)\; \label{line:uncertainty}
  }{
    EVALUATE(random $x_\text{TXF}, x_\text{ED}$) \Comment*{Diversity sampling} \label{line:diversity}
  }
  }
  $\delta \gets \delta \ \cup$ \{new performance point ($x_\text{TXF}, x_\text{ED}, P$)\}; \\ 
 }
 \caption{BOSHCODE} 
 \label{alg:boshcode}
\end{algorithm}

Algorithm~\ref{alg:boshcode} summarizes the above process. Starting from an initial sample set $\delta$, we run until convergence the following steps. To trade off between exploration and exploitation, we consider two probabilities: uncertainty-based exploration ($\alpha_P$) 
and diversity-based exploration ($\beta_P$). With probability $1 - \alpha_P - \beta_P$, we run second-order GOBI using the surrogate model to optimize UCB. Adding the converged point 
$(x, o)$ in $\delta$, we train the surrogate models (line~\ref{line:fit} in Algorithm~\ref{alg:boshcode}). We then generate a new query point (using GOBI), transfer weights from neighboring models, and 
train it (or use a pre-trained surrogate) through the EVALUATE function (lines~\ref{line:opt}-\ref{line:eval}). With $\alpha_P$ probability, we sample the search space using 
the combination of aleatoric and epistemic uncertainties to find a 
point where the performance estimate is uncertain (line~\ref{line:uncertainty}). To avoid getting stuck in a 
localized search subset, we also choose a random point with probability $\beta_P$ (line~\ref{line:diversity}). The EVALUATE function gives the performance measure $P$ for the given pair $(x_\text{TXF}, x_\text{ED})$ using the ProTran and FlexiBERT 2.0 frameworks or their corresponding surrogates.

\subsection{The GPTran Framework}
\label{sec:gptran_framework}

Once EdgeTran obtains the best pair of transformer model and edge device, GPTran nudges the architectural 
parameters of the converged transformer architecture to improve performance further. Unlike BOSHCODE, which 
globally searches for the best-performing architecture, GPTran is a local search post-processing technique. 
Its operation takes inspiration from the lottery ticket hypothesis~\cite{lottery_ticket}, where a part of the 
network is usually sufficient to obtain the same performance as the parent network. In our case, GPTran also 
helps overcome inaccuracies in surrogate modeling that may lead the co-design framework to a model that is 
close to but not precisely optimal. However, unlike previous works on structural adaptation at the level of 
individual neurons or convolutional filters (for CNNs)~\cite{nest}, we run our grow-and-prune framework at the 
compute-block level due to the modularity of the FlexiBERT 2.0 design space (details in 
Section~\ref{sec:flexibert_2}).

GPTran runs gradient-based (along with random) growth and magnitude-based pruning at the block level for 
transformer architectures. It involves multiple iterations of interleaved grow-and-prune steps. We describe
these steps next.

\begin{itemize}
    \item \textbf{The grow step:} For a given parent model, $n_\text{G}$ child models are instantiated with 
the net number of parameters slightly higher than that of the parent. Here, we employ either of two types of 
growth strategies:
    \begin{itemize}
        \item \underline{Grow attention head ($G_\text{A}$)}: We add an attention head (chosen from a set 
$A_\text{TXF}$) to a particular encoder based on two scenarios with equal probability. We either add an attention head next to the one with the highest (or the next highest) gradient or at random. Here, we add $n^\text{G}_\text{A}$ operation blocks. As expected, we also increase the hidden dimension $h^j$ for the selected layer by the hidden 
dimension of the added attention head since the net hidden dimension of an encoder layer is a sum of those for each attention head.
        \item \underline{Grow feed-forward stack ($G_\text{FF}$)}: We add a fully-connected layer to the 
stack, in the feed-forward module, with $\textbf{min}(h_\text{F}, h^\text{G}_\text{F})$ neurons, where 
$h_\text{F}$ is the number of neurons in the last hidden layer in the selected feed-forward module and 
$h^\text{G}_\text{F}$ is a predetermined hyperparameter. Again, we select the feed-forward module based on 
the gradient or randomly, each with equal probability.
    \end{itemize}
    We generate all the $n_\text{G}$ children based on the current growth mode (either $G_\text{A}$ or 
$G_\text{FF}$).
    \item \textbf{The prune step:} For a given parent model, we instantiate a child model (number of children 
$n_\text{P} = 1$) with the net number of parameters slightly lower than that of the parent. For this, we 
employ either of the following two pruning strategies:
    \begin{itemize}
        \item \underline{Prune attention head ($P_\text{A}$)}: We remove $n^\text{P}_\text{A}$ attention heads 
based on their average magnitude of weights. For instance, for a WMA head, we obtain an average of all weight 
matrices (i.e., key, query, value, output, and the WMA matrices). If the average of the weights for this head 
is the lowest among all heads in the current model, we prune it out from the child model (which was initially a replica of the parent). 
Again, we also reduce the hidden dimension $h^j$ of the selected layers by the hidden dimensions of the 
attention heads removed.
        \item \underline{Prune feed-forward layer ($P_\text{FF}$)}: We prune a fully-connected layer based on 
the average weights of the fully-connected layers in all feed-forward modules. We prune the selected layer to 
$\textbf{min}(h^\text{P}_\text{F}, h_\text{F} - h^\text{P}_\text{F})$ number of neurons, where $h_\text{F}$ 
is the number of neurons in the selected hidden layer and $h^\text{P}_\text{F}$ is a predetermined 
hyperparameter.
    \end{itemize}
    We prune the selected model based on the current pruning mode (either $P_\text{A}$ or $P_\text{FF}$) 
employing either of the above strategies.
\end{itemize}

To search for compact models from the current converged model obtained using BOSHCODE, we set $n^\text{P}_\text{A}$
to be higher than $n^\text{G}_\text{A}$ (more details in Section~\ref{sec:exp_setup_gp}). To minimize the
number of training steps for every child node and leverage the neighboring (and already trained) parent
node, we transfer the weights via the RP or OT method described in Section~\ref{sec:weight_transfer}. Due to 
the high overlap ratio (between the parent and the child) and highly granular weight transfer in FlexiBERT 2.0, we can train individual child 
models rapidly. This significantly reduces search time.

For GPTran, the optimization metric is the pre-training loss. Unlike some previous
works~\cite{movement_pruning, rethinking}, we employ block-level growth and pruning during pre-training rather than during
fine-tuning~\cite{compressing_bert}. Thus, the optimization metric is the pre-training loss (or the model's 
perplexity on language data) while executing local search. We implement GPTran in a cycle of four \emph{modes} 
($M_\text{GP}$) in the following order: $G_\text{A}$, $G_\text{FF}$, $P_\text{A}$, $P_\text{FF}$. We cycle through the 
grow/prune modes at every tree depth until we reach the best-performing architecture (i.e., one whose children perform 
worse than that node).

\SetKwComment{Comment}{/* }{ */}
\SetKw{KwInFor}{\textbf{in}}
\begin{algorithm}[t]
\SetAlgoLined
\KwResult{optimal transformer model}
 \textbf{Initialize:} root-node model, best-node $\gets$ root-node, modes $M_\text{GP}$, $i \gets 1$; \\
 \While{children\textup{(best-node)} $<$ \textup{best-node}}{
  $i$ $\gets$ ($i$ + 1) $\bmod$ $len(M_\text{GP}) + 1$; \\
  \uIf{$M_\textup{GP}[i] = G_\textup{A}$}{ 
    \For{$j \gets 1$ \KwTo $n^\textup{G}_\textup{A}$}{
     child $\gets$ best-node; \\
     \eIf{prob $\sim U(0,1) < \alpha^\textup{G}_\textup{A}$}{
      $a_\text{GRAD} \gets \textbf{argmax}(\nabla a, \ \forall a \in \textup{child})$; \label{line:attn_grad} \\
      child $\gets$ child + $a_\text{GRAD} \in A_\text{TXF}$; 
     }{
     child $\gets$ child + random $a_\text{TXF} \in A_\text{TXF}$; \label{line:attn_rand} \\
     }
    }
  }
  \uElseIf{$M_\textup{GP}[i] = G_\textup{FF}$}{
    \For{$j \gets 1$ \KwTo $n^\textup{G}_\textup{A}$}{
     child $\gets$ best-node; \\
     \eIf{prob $\sim U(0,1) < \alpha^\textup{G}_\textup{A}$}{
      $f_\text{GRAD} \gets \textbf{argmax}(\nabla f, \ \forall f \in \textup{child})$; \label{line:ff_grad} \\
      $f_\text{GRAD} \gets f_\text{GRAD} + f_\text{I}(\textbf{min}(h_{f_\text{GRAD}}, h^\text{G}_\text{F}))$; \\
     }{
     select random $f \in \text{child}$; \label{line:ff_rand} \\
     $f \gets f + f_\text{I}(\textbf{min}(h_{f}, h^\text{G}_\text{F}))$; \\
     }
    }
  }
  \uElseIf{$M_\textup{GP}[i] = P_\textup{A}$}{
    child $\gets$ best-node; \\
    $a_\text{S} \gets sort(a, \ \forall a \in \text{child})$; \\
    \For{$j \gets 1$ \KwTo $n^\textup{P}_\textup{A}$}{
     child $\gets$ child $- \ a_\text{S}[j]$; \label{line:attn_mag} \\
    }
  }
  \uElseIf{$M_\textup{GP}[i] = P_\textup{FF}$}{
    child $\gets$ best-node; \\
    $f_\text{S} \gets sort(f, \ \forall f \in \text{child})$; \\
    \For{$j \gets 1$ \KwTo $n^\textup{P}_\textup{A}$}{
     $f_\text{S}[j] \gets f_\text{I}(\textbf{min}(h^\text{P}_\text{F}, h_\text{F} - h^\text{P}_\text{F})$); \label{line:ff_mag} \\
    }
  }
  \For{\textup{child} \KwInFor children\textup{(best-node)}}{
   $W_\text{child} \gets W_\text{best-node}$; \label{line:weight_transfer} \\
   train child; \\
  }
  best-node $\gets$ child with minimum loss; \\
 }
 \caption{GPTran} 
 \label{alg:gptrain}
\end{algorithm}

Algorithm~\ref{alg:gptrain} summarizes the GPTran algorithm. It stops at the best-performing model. It starts 
with the converged transformer model obtained using BOSHCODE. Then, it cycles through the four modes presented 
above. For $G_\text{A}$, it creates $n^\text{G}_\text{A}$ child models based on the attention head with the 
maximum gradient during training (line~\ref{line:attn_grad}) or a randomly selected attention head 
(line~\ref{line:attn_rand}). For $G_\text{FF}$, it grows a feed-forward stack based on the one with the highest 
gradient during training (line~\ref{line:ff_grad}) or a randomly selected layer (line~\ref{line:ff_rand}). 
Here, function $f_\text{I}()$ refers to the instantiation of a new fully-connected layer with the hidden 
dimension as input. For $P_\text{A}$, it removes the attention heads with the $n^\text{P}_\text{A}$ smallest average 
weight magnitudes. For $P_\text{FF}$, it prunes feed-forward layers with $n^\text{P}_\text{A}$ smallest 
average weight magnitudes by a given factor. Finally, we transfer weights from the parent node 
to the instantiated children before training them (line~\ref{line:weight_transfer}). Here, we implement 
weight transfer through OT or RP (see Section~\ref{sec:flexibert_2}). Note that we randomly instantiate the 
weights of all newly added attention heads or layers.

GPTran also implements backtracking~\cite{backtracking} (not shown in Algorithm~\ref{alg:gptrain}) when a 
current best-performing leaf node does not give the overall best pre-training loss. In the hierarchical 
\emph{tree} data structure formed during search, if the currently reached leaf does not have the 
best performance (or the lowest pre-training loss), GPTran backtracks to the node with the next-best 
performance that has unexplored children. It then populates the tree from there.

\section{Experimental Setup}
\label{sec:exp_setup}

This section presents the setup for various experiments we perform, along with the baselines for comparison.

\subsection{FlexiBERT 2.0 Design Space}
\label{sec:exp_setup_flexibert_2}

Table~\ref{tab:hyp_ranges} shows the range of hyperparameter values in the proposed FlexiBERT 2.0 design space. 
This expanded range for each hyperparameter increases the number of possible transformer models from 
3.3 $\times$ 10$^9$ in the original FlexiBERT framework to 1.7 $\times$ 10$^{88}$. A large 
design space leads to better-performing models, which motivates this expansion~\cite{flexibert, codebench}. 

\subsubsection{Hyperparameter Combinations}

Next, we illustrate the process of obtaining the many architectures in our design space.

\begin{itemize}
    \item Different feed-forward hidden dimensions are possible (6 values as per Table~\ref{tab:hyp_ranges}). We can stack these feed-forward operations with 1, 2, or 3 hidden layers. Thus, the number of feed-forward operation types $= \ 6 + 6^2 + 6^3 =$ 258.
    \item There are 7 possible attention operations in $A_\text{TXF}$, namely: SA-SDP, SA-WMA, LT-DFT,
LT-DCT, DSC-5, DSC-9, and DSC-13. Thus, the number of multi-head attention operation types possible for
each encoder layer (without considering the hidden dimension) $ = \sum_{i \in n_\text{A}}{7+i-1 \choose
i}$ (for $n_\text{A} = \{$2, 4, 6, 8, 10, 12$\}$ attention heads each) = 21805. Note that we have used 
combinations with replacement, i.e., $n+i-1 \choose i$, and not product, i.e., $n^i$, since that would add 
isomorphic redundancy to every encoder layer.
    \item Now, for every encoder layer, we need to determine the feed-forward operations, hidden dimension, 
and multi-head attention operation, leading to $\sum_{i \in n_\text{A}} 258^i \times 4^i \times 21805^i = $ 
1.7 $\times$ 10$^{88}$ transformer models.
\end{itemize}

\subsubsection{Model Training}

We pre-train our models with a combination of publicly available text corpora, viz.~\texttt{BookCorpus} 
(BookC) \cite{bookcorpus}, \texttt{Wikipedia English} (Wiki), \texttt{OpenWebText} (OWT) \cite{OpenWebtext}, 
and \texttt{CC-News} (CCN) \cite{CC_news}. We borrow most training hyperparameters from RoBERTa~\cite{roberta} 
for robust training of diverse architectures in our design space. We set the batch size to 256 and
warm up the learning rate over the first $10,000$ steps to its peak value at $1 \times 10^{-4}$ that then decays 
linearly. We set the weight decay to $0.01$, Adam scheduler's parameters $\beta_1 = 0.9$, $\beta_2 = 0.98$ 
(shown to improve stability; \cite{roberta}), $\epsilon = 1 \times 10^{-6}$, and run pre-training for 
$1,000,000$ steps. 

We fine-tune our models on the nine GLUE tasks~\cite{glue}. We also run automatic hyperparameter tuning in
the fine-tuning process (i.e., search the training recipe) using the tree-structured Parzen estimator algorithm 
\cite{optuna_2019}. We randomly select the learning rate logarithmically in the [$2\times 10^{-5},\;5 \times 
10^{-4}$] range and batch size in $\{16, 32, 64\}$ uniformly. Table~\ref{tab:trancode_training_recipe} shows the 
best training recipe for fine-tuning ET (\underline{e}dge device and \underline{t}ransformer co-design model we obtain from BOSHCODE) on each GLUE task selected using 
this autotuning technique. This hyperparameter optimization uses random initialization each time. This results 
in variation in performance each time we query the model (otherwise called the aleatoric uncertainty). For 
tasks MRPC, RTE, and STS-B, we use the fine-tuned checkpoint from MNLI training instead of the pre-trained 
model~\cite{roberta}

We train all models on NVIDIA A100 GPUs and 2.6 GHz AMD EPYC Rome processors. 
The entire process of training the 16 LHS samples in the FlexiBERT 2.0 design space took around 100 GPU-days.

\begin{table}
\caption{Hyperparameters used for fine-tuning ET on the GLUE tasks.}
\centering
\begin{tabular}{@{\hskip 0.2in}l@{\hskip 0.5in}c@{\hskip 0.5in}c@{\hskip 0.2in}}
\toprule
Task & Learning rate & Batch size \\ \midrule
CoLA & $2.0 \times 10^{-4}$ & 64 \\ [1mm]
MNLI & $9.4 \times 10^{-5}$ & 64 \\ [1mm]
MRPC & $2.23 \times 10^{-5}$ & 32 \\ [1mm]
QNLI & $5.03 \times 10^{-5}$ & 128 \\ [1mm]
QQP & $3.7 \times 10^{-4}$ & 64 \\ [1mm]
RTE & $1.9 \times 10^{-4}$ & 128 \\ [1mm]
SST-2 & $1.2 \times 10^{-4}$ & 128 \\ [1mm]
STS-B & $7.0 \times 10^{-5}$ & 32 \\ [1mm]
WNLI & $4.0 \times 10^{-5}$ & 128 \\ \bottomrule
\end{tabular}
\label{tab:trancode_training_recipe}
\end{table}

\subsubsection{Surrogate Modeling}
\label{sec:exp_setup_surrogate_modeling}

To obtain a surrogate model for all transformer architectures in the FlexiBERT 2.0 design space, we employ a 
similar approach to ProTran. For the initial 16 LHS samples, we test different regressors, as described in 
Section~\ref{sec:learning_perf_pred}. However, while co-designing without hard constraints on model accuracy, 
a user may be interested in the \emph{best} model from a sampled set instead of the one that barely meets 
the accuracy constraint. For this, we also test a ranking regressor, LambdaMART, which represents a state-of-the-art in the learning-to-rank problem~\cite{lambda_mart}.

\begin{figure}
    \centering
    \includegraphics[width=\linewidth]{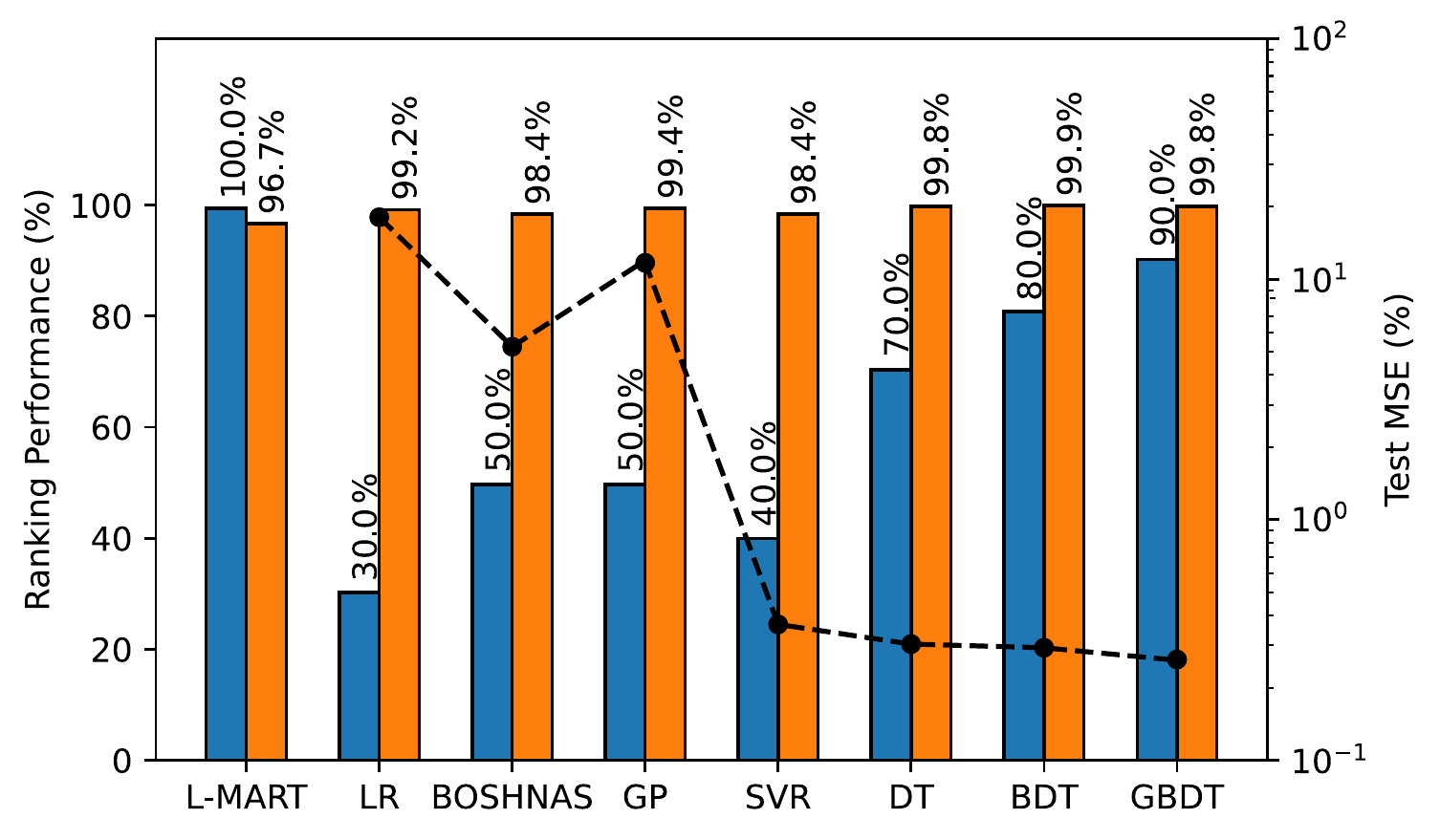}
    \caption{Surrogate modeling performance in terms of (a) ranking performance of the `best' (blue) and 
nDCG (orange) ranking tests on the left, and (b) test MSE on the GLUE score predictions on the right. 
Absolute test MSE for L-MART is not shown since it is only a relative ranking model.}
    \label{fig:surrogate_models_flexibert}
\end{figure}

Fig.~\ref{fig:surrogate_models_flexibert} compares different regressors based on ranking performance and MSE on 
the test set for the prediction of GLUE scores. We take the first 11 models in the initial set of LHS samples as 
the training set and measure performance on the rest, i.e., the test set. We compare ranking performance based 
on two tests. First, we assemble all models in the training set into groups of three (resulting in 
${5 \choose 3}$ combinations in the test set). Then, we compare the best model by taking the actual best model 
among the set of five and the predicted best model by sorting the models based on their predicted GLUE scores. 
We also compare a commonly used ranking metric, the normalized discounted cumulative gain (nDCG)~\cite{ndcg}. 
For this, we discount subsequent ranks logarithmically. Finally, we compare the absolute MSE for the predicted 
GLUE scores on the test set. Although LambdaMART (L-MART) has a high ranking performance, GBDT shows reasonably high 
ranking performance and low test MSE (0.3\%). Hence, we use the GBDT surrogate model for performance prediction 
in our design space.

\subsection{Hardware Performance Measurements}

\subsubsection{Hardware Platforms}
\label{sec:exp_setup_hw_platforms}

We now present details of the hardware platforms that form the test bed for our experiments. The baseline 
server platform we consider is the Nvidia A100 GPU with 40 GB video random-access memory (VRAM). The mobile platforms 
include the Apple M1 ARM SoC~\cite{apple_m1} with an 8-core CPU, 8-core GPU, and 16 GB unified memory on an iPad 
(for ease of experimentation, we instead perform experiments on a MacBook Pro that has the same SoC), Raspberry Pi 
4 Model-B~\cite{rpi} that has the Broadcom BCM2711 ARM SoC, Intel Neural Compute Stick v2 with its NPU~\cite{ncs}, 
and, finally, an Nvidia Jetson Nano~\cite{nano} with an Nvidia Tegra X1 ARM SoC that has a CPU, an embedded GPU, and 
2 GB unified memory.

\subsubsection{Power Measurements}
\label{sec:exp_setup_power_meas}

We use an INA219 sensor, connected via the I2C interface to Raspberry Pi, for energy and power measurements of the Raspberry Pi, Nvidia Jetson Nano, and the Intel Neural Compute Stick. \textcolor{black}{The sensor measures real-time power drawn by the device power supply. Thus, the measurement corresponds to the net energy consumed by the hardware platform. For the Nvidia A100 GPU, we use the \texttt{nvidia-smi} command to measure GPU power draw. For the Apple M1 processor, we measure the CPU/GPU power via the \texttt{powermetrics} command. These commands measure the power drawn by the power supply of the respective hardware modules.} 

We perform all measurements of hardware performance while running inference on the GLUE tasks (multiple times). We then take the geometric mean of the evaluated performance measures and train the surrogate models with these mean profiles.


\subsection{Co-design Pipeline}

\begin{table}
\caption{Hyperparameters used in GPTran.}
\centering
\begin{tabular}{@{\hskip 0.2in}cc@{\hskip 0.2in}}
\toprule
Hyperparameter                  & Value(s)                       \\ \midrule
\multirow{2}{*}{$A_\text{TXF}$} & SA-SDP, SA-WMA, LT-DFT, LT-DCT \\ [1mm]
                                & DSC-5, DSC-9, DSC-13            \\ [1mm]
$n_\text{G}$                    & 10                              \\ [1mm]
$n^\text{G}_\text{A}$           & 1                              \\ [1mm]
$h^\text{G}_\text{F}$           & 1024                           \\ [1mm]
$n_\text{P}$                    & 1                              \\ [1mm]
$n^\text{P}_\text{A}$           & 2                              \\ [1mm]
$h^\text{P}_\text{F}$           & 128                            \\ \bottomrule
\end{tabular}
\label{tab:gptran_hyperparameters}
\end{table}

To run BOSHCODE, we use the following parameter values to obtain the net performance measure: $\alpha = 0.5$, 
$\beta = 0.2$, $\gamma = 0.2$, $\epsilon = 0.1$ (Eq.~\ref{eqn:perf_metric}). We set $\alpha_P$ and $\beta_P$ to 
0.1 each, and $k_1$ and $k_2$ to 0.5 each. For all three surrogate models $f$, $g$, and $h$, we pass the input 
embeddings of the transformer model and edge device ($x_\text{TXF}$ and $x_\text{ED}$, respectively) to networks 
with two distinct hidden layers with 32 hidden neurons each. We then concatenate the outputs of these two separate 
sub-networks and pass them through a fully-connected layer with 64 and then 32 neurons. Finally, the network ends 
with one output neuron to predict the performance measure.

All input embeddings obtained using GOBI for the surrogate models may not be valid. For instance, $x_\text{ED}$ 
should be one-hot encoded. To add constraints to the optimization process, along with forcing the model to learn 
the performance only for valid input embeddings, we add a datapoint ($x_\text{TXF}$, $x_\text{ED}$, $P_\text{MIN}$) 
to the dataset $\delta$ if either of the input embeddings is invalid or does not adhere to input constraints. 
Another example of input constraint could be that transformers with only up to six layers are allowed. 
$P_\text{MIN}$ has a very low value, set to $-$100 for our experiments.

\subsection{Grow-and-Prune Process Applied to ET}
\label{sec:exp_setup_gp}

We apply GPTran to the optimal transformer model, i.e., ET, produced by BOSHCODE. 
Table~\ref{tab:gptran_hyperparameters} summarizes the hyperparameters chosen for GPTran. 
Table~\ref{tab:gptran_training} shows the training choices for each of the four modes ($M_\text{GP}$ 
described in Section~\ref{sec:gptran_framework}). We found all hyperparameter values through grid search.

\subsection{Baselines}
\label{sec:baselines}

Our baseline models include BERT-Base, a hand-designed transformer model. We also include a model obtained
through a throughput-guided NAS technique (AutoTinyBERT~\cite{autotinybert}). However, it only relies on throughput 
measurements on a CPU. HAT~\cite{hat_mit}, another baseline for comparison, uses latency feedback from selected 
devices for a guided search. It runs hardware-aware NAS (HW-NAS), which searches for transformer models with 
latency feedback from a given hardware platform. This, however, loses the benefits of co-design in a context where 
multiple edge devices may be employed. For fair comparisons, we present an HW-NAS version of EdgeTran in which we 
run BOSHCODE but force gradients to the edge device to zero, i.e., we only search for transformer models run on a 
given edge platform (here, a Raspberry Pi).

\begin{table}
\caption{Training choices for different modes in GPTran.}
\centering
\begin{tabular}{@{\hskip 0.2in}c@{\hskip 0.2in}c@{\hskip 0.2in}c@{\hskip 0.2in}}
\toprule
Mode & Max. Learning Rate & Pre-training Steps \\ \midrule
$G_\text{A}$ & $1 \times 10^{-5}$ & 20,000 \\ [1mm]
$G_\text{FF}$ & $1 \times 10^{-5}$ & 20,000 \\ [1mm]
$P_\text{A}$ & $5 \times 10^{-5}$ & 20,000 \\ [1mm]
$P_\text{FF}$ & $1 \times 10^{-5}$ & 10,000 \\ \bottomrule
\end{tabular}
\label{tab:gptran_training}
\end{table}

\begin{figure*}[t]
    \centering
    \includegraphics[width=\linewidth]{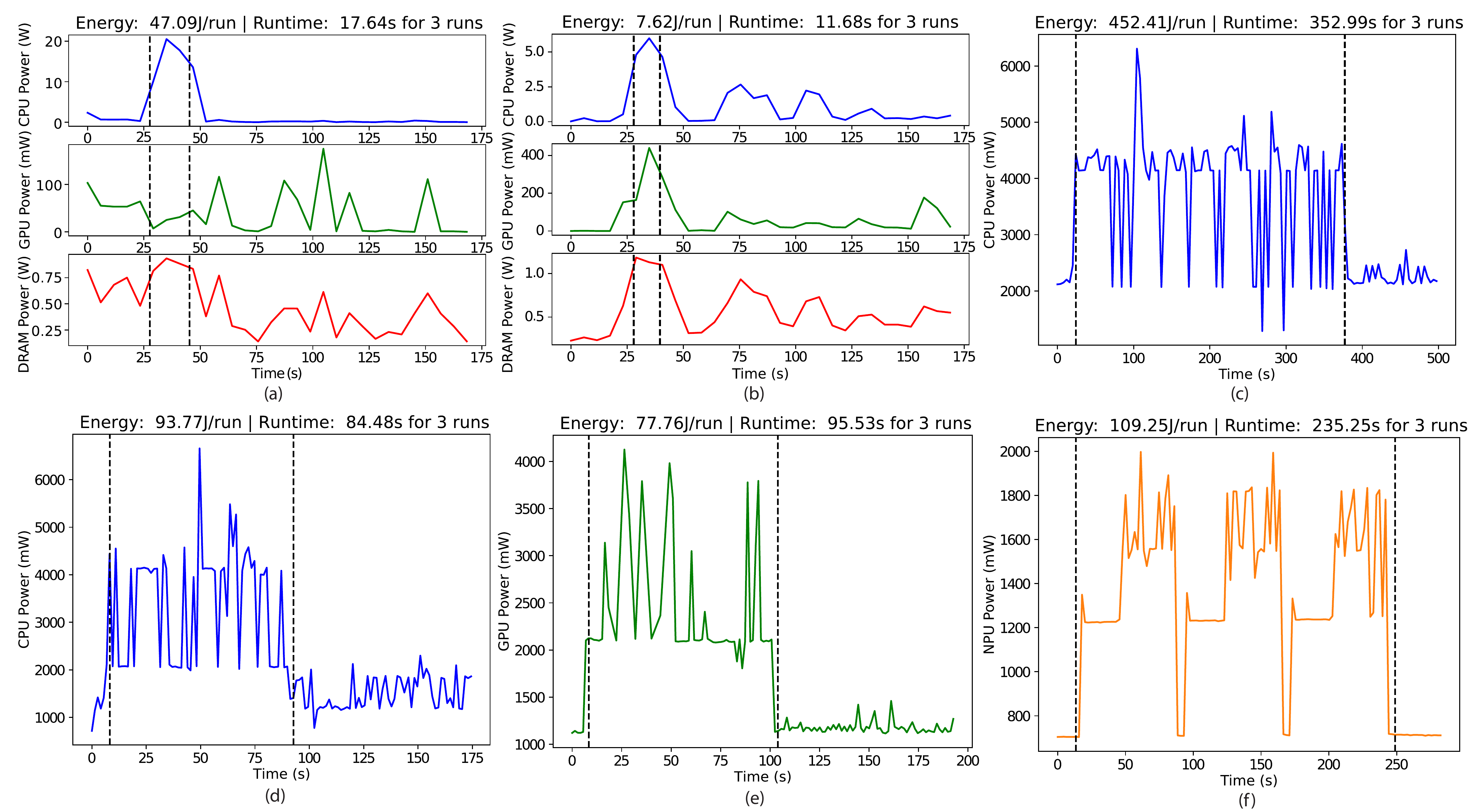}
    \caption{Power consumption from different sources (CPU, GPU, or DRAM) for different platforms: (a) Apple M1 CPU, 
(b) Apple M1 GPU, (c) Raspberry Pi CPU, (d) Nvidia Jetson Nano CPU, (e) Nvidia Jetson Nano GPU, and (f) Intel Neural 
Compute Stick NPU. One run corresponds to a full pass of running inference of the BERT-Tiny~\cite{turc2019} model 
on the SST-2~\cite{glue} task for the entire dataset.}
    \label{fig:power_consumption}
\end{figure*}

\section{Results}
\label{sec:results}

This section presents experimental results and comparisons of the EdgeTran framework with relevant baselines.

\subsection{Hardware Performance Comparisons}

We now compare the hardware performance measures, namely latency, energy consumption, and peak power draw, 
on different platforms, for a given transformer model. This demonstrates the capabilities of each hardware 
platform with like-for-like comparisons.

\begin{figure*}[t]
    \centering
    \includegraphics[width=\linewidth]{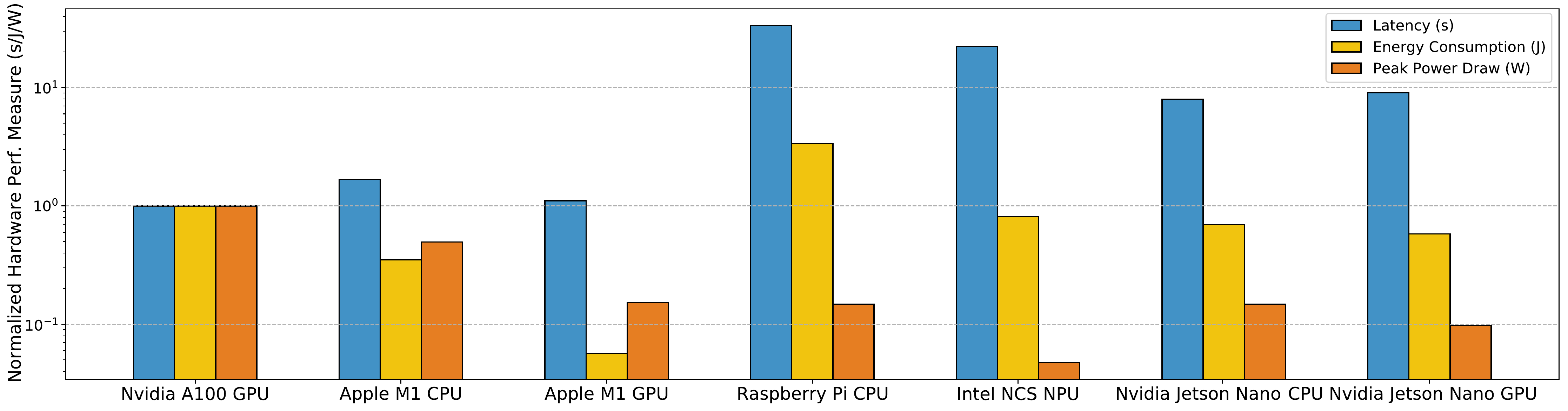}
    \caption{Gains in different hardware performance measures for diverse embedded platforms. Results have been normalized against those for the baseline, i.e., the Nvidia A100 GPU.}
    \label{fig:protran_gains}
\end{figure*}

Fig.~\ref{fig:power_consumption} shows power consumption while running model inference with 
BERT-Tiny~\cite{turc2019} on the SST-2 task~\cite{glue} for different hardware platforms. 
Figs.~\ref{fig:power_consumption}(a) and (b) show the CPU, GPU, and dynamic random access memory (DRAM) power 
consumption for the Apple M1 SoC~\cite{apple_m1}. Fig.~\ref{fig:power_consumption}(c) shows power consumption for 
the Raspberry Pi CPU. Figs.~\ref{fig:power_consumption}(d) and (e) show power consumption for Nvidia Jetson Nano 
when we run the model on the CPU and GPU, respectively. Finally, Fig.~\ref{fig:power_consumption}(f) shows power 
consumption for Intel NCS v2. These figures show that the mobile platforms have much lower power consumption 
throughout their operation when compared to the A100 GPU, which has a peak power draw of around 42W. These profiles 
also highlight the diverse power draw characteristics and peak power consumption of different platforms. ProTran 
automatically profiles these curves while running a search in its active-learning pipeline.

\begin{figure*}[ht]
    \centering
    \includegraphics[width=\linewidth]{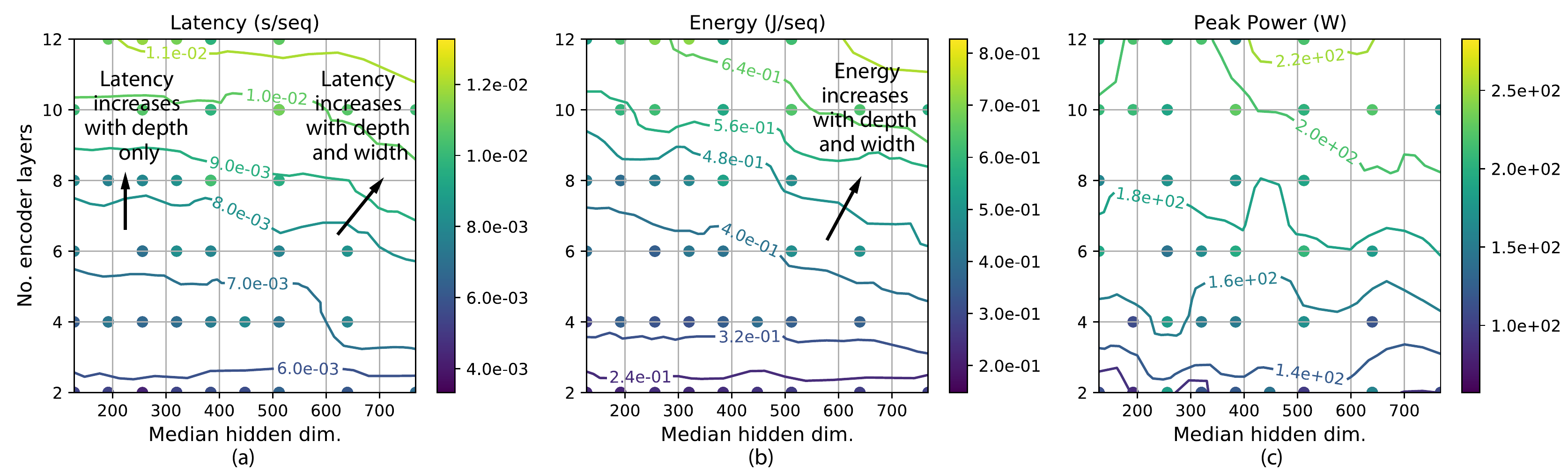}
    \caption{Profiles of hardware performance measures on the Nvidia A100 GPU for transformer model depth and 
width: (a)~latency, (b) energy, and (c) peak power.}
    \label{fig:contour_a100}
\end{figure*}

As can be seen from Fig.~\ref{fig:protran_gains}, the Apple M1 SoC with its integrated GPU outperforms the traditional 
A100 GPU in terms of energy and peak power consumption while being close in terms of latency. Its total energy 
consumption is the minimum among all platforms under consideration. We see a 17.6$\times$ reduction in energy 
consumption and 6.6$\times$ reduction in peak power draw with only a 10.6\% higher latency per run for the Apple 
M1 SoC running on its integrated GPU when compared to the Nvidia A100 GPU. On the other hand, Intel NCS has the 
minimum peak power consumption, i.e., with a 21.0$\times$ reduction, but with a 22.3$\times$ higher latency 
compared to the Nvidia A100 GPU. This demonstrates a diverse set of latency, energy, and peak power draw profiles 
for the different platforms that might fit various requirements or constraints on hardware performance for assorted 
edge-AI deployments. This diversity motivates the need for profiling the latency, energy, and peak power consumption 
of various transformer architectures on a diverse set of embedded platforms. This would lead to efficient 
transformer architectural decisions under different scenarios.

\subsection{Effect of Model size on Hardware Performance}

Fig.~\ref{fig:contour_a100} shows contour plots for latency, energy consumption, and peak power draw when running inference on the Nvidia A100
GPU. The plots show the dependence of these hardware performance measures on model depth and width specified in 
terms of the number of encoder layers and the median hidden dimension of all encoder layers, respectively. The color of the point 
plots depicts the performance value. We obtain the contour 
plots from a trained surrogate for each performance measure. As can be seen 
from Fig.~\ref{fig:contour_a100}(a), latency is highly dependent on model depth and does not have a high correlation 
with model width (based on the horizontal contour lines). The sequential operation of the encoder layers in a transformer model is the reason
behind this result. Thus, as long as the target device runs an entire layer in parallel, the latency should increase linearly with 
the number of encoder layers. However, for deeper models (i.e., with the number of encoder layers more than 2), latency increases slightly with model width as well (as the contour lines are not horizontal anymore). For instance, the typical latency of a model with 10 encoder layers and a median hidden dimension of 512 is less than $1 \times 10^{-2}$ s/seq. while that of a model with the same number of layers but a hidden dimension of 640 is higher than $1 \times 10^{-2}$ s/seq. 
Fig.~\ref{fig:contour_a100}(b) shows the energy consumption profiles with model depth and width. The net energy 
consumption increases with the depth and width of the transformer model. We explain this as follows. 
As the model depth and width increase, the number of computations increases, thus raising the hardware energy 
consumption for these computations. Peak power profiles in Fig.~\ref{fig:contour_a100}(c) behave in a much more 
convoluted fashion when model depth and width are varied. We observe trenches in peak power profiles for certain 
hidden dimensions. This could be attributable to how well the model fits in GPU memory for different hidden 
dimensions. Nevertheless, there is still a general trend of increasing peak power consumption with model depth. 

\begin{table*}[]
\caption{Hardware performance measures of optimal models in terms of latency, energy, and peak power on different platforms.}
\centering
\begin{tabular}{@{}l|ccc|ccc|ccc@{}}
\toprule
\multirow{2}{*}{\textbf{Device}} & \multicolumn{3}{c|}{\textbf{Min. Latency Model}} & \multicolumn{3}{c|}{\textbf{Min. Energy Model}} & \multicolumn{3}{c}{\textbf{Min. Peak Power Model}} \\ [1mm] \cmidrule(l){2-10} 
 & \begin{tabular}[c]{@{}c@{}}Latency \\ (ms/seq)\end{tabular} & \begin{tabular}[c]{@{}c@{}}Energy \\ (J/seq)\end{tabular} & \begin{tabular}[c]{@{}c@{}}Peak Power \\ (W)\end{tabular} & \begin{tabular}[c]{@{}c@{}}Latency \\ (ms/seq)\end{tabular} & \begin{tabular}[c]{@{}c@{}}Energy \\ (J/seq)\end{tabular} & \begin{tabular}[c]{@{}c@{}}Peak Power \\ (W)\end{tabular} & \begin{tabular}[c]{@{}c@{}}Latency \\ (ms/seq)\end{tabular} & \begin{tabular}[c]{@{}c@{}}Energy \\ (J/seq)\end{tabular} & \begin{tabular}[c]{@{}c@{}}Peak Power \\ (W)\end{tabular} \\ [1mm] \midrule
Nvidia A100 GPU & 5.69 & 0.25 & 133.94 & 5.95 & 0.24 & 133.94 & 5.95 & 0.24 & 98.04 \\ [1mm]
Apple M1 CPU & 159.6 & 3.14 & 23.24 & 159.6 & 3.13 & 22.21 & 652.3 & 12.06 & 21.76 \\ [1mm]
Apple M1 GPU & \textbf{4.35} & 0.065 & 18.18 & 6.56 & \textbf{0.064} & 18.40 & 6.54 & 0.065 & 18.18 \\ [1mm]
Raspberry Pi CPU & 2113.2 & 28.09 & 4.48 & 6417.4 & 28.09 & 4.48 & 16843.4 & 52.27 & 4.44 \\ [1mm]
Intel NCS NPU & 5263.8 & 6.73 & 2.12 & 5741.2 & 6.21 & 2.08 & 65576.5 & 81.05 & \textbf{2.01} \\ [1mm]
Jetson Nano CPU & 6562.6 & 15.03 & 4.08 & 8726.6 & 15.01 & 4.08 & 7787.1 & 20.06 & 4.06 \\ [1mm]
Jetson Nano GPU & 41425.7 & 92.19 & 4.14 & 42365.2 & 81.27 & 4.10 & 128793.4 & 309.4 & 4.02 \\ \bottomrule
\end{tabular}
\label{tab:opt_measures}
\end{table*}

\begin{figure*}
    \centering
    \includegraphics[width=\linewidth]{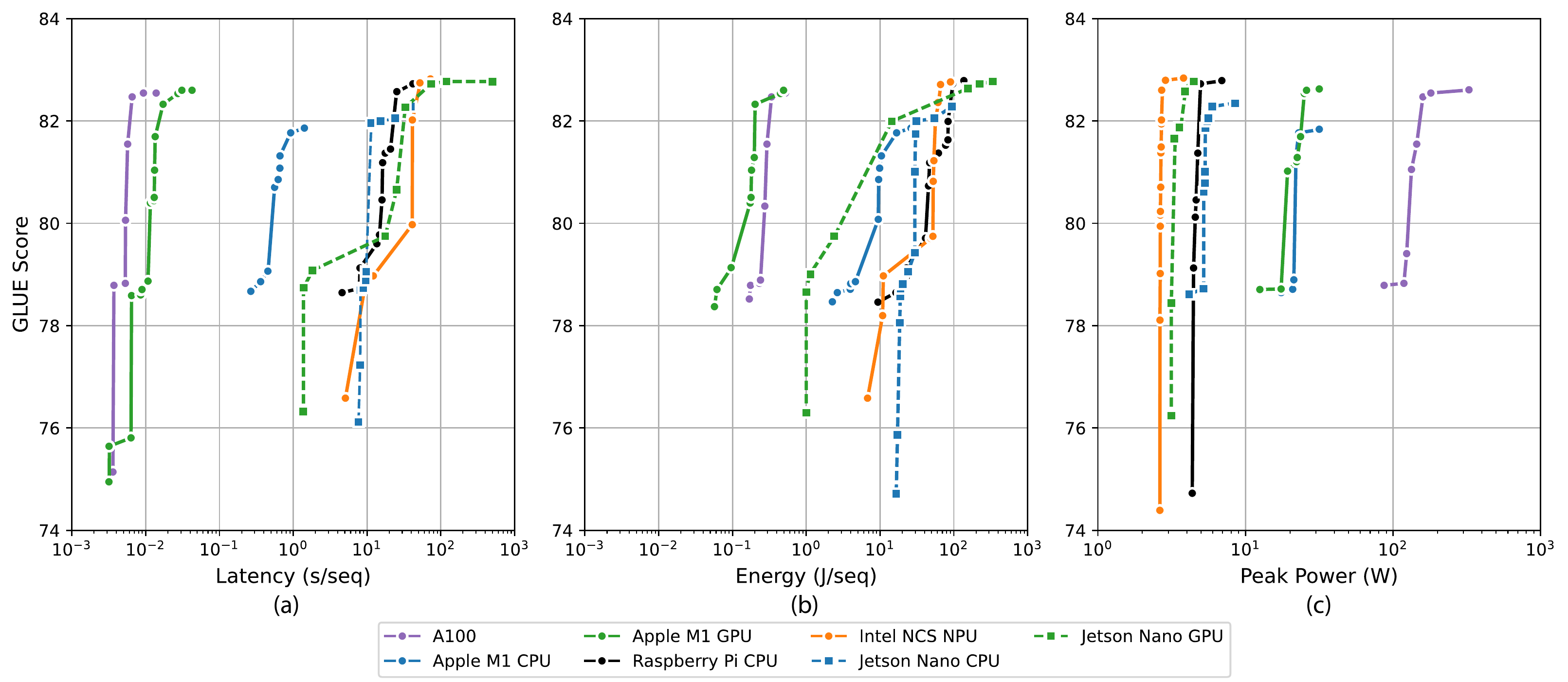}
    \caption{Pareto frontiers of GLUE scores for models in the FlexiBERT 2.0 design space with (a) latency, 
(b) energy, and (c)~peak power for different platforms.}
    \label{fig:pareto}
\end{figure*}

Table~\ref{tab:opt_measures} presents the best models for minimizing latency, energy, and peak power draw 
for each hardware platform. We observe that relative to the best-latency model running on Nvidia A100 GPU, the 
best-latency model running on Apple M1 GPU incurs 22.0\% lower latency, 3.8$\times$ lower energy consumption, and 
7.4$\times$ lower peak power draw. The best-energy model running on Apple M1 GPU also has the lowest energy 
among all such models, achieving 3.7$\times$ lower energy with only 10.2\% higher latency but 7.3$\times$ lower 
peak power draw relative to the A100 GPU. Other platforms have higher energy consumption due to drastically higher 
latencies. The Intel NCS NPU has the lowest peak power draw, 48.8$\times$ lower than that of the A100 GPU, 
however, with 11,021.3$\times$ higher latency and 337.7$\times$ higher energy consumption. Nevertheless, in edge 
applications with a restriction on the input power supply to 2W, the Intel NCS might be the designer's best bet.

\subsection{Pareto Frontiers}

The above comparisons do not consider the effect of transformer model size (and corresponding design decisions) on accuracy when differentiating 
one platform from another. Hence, we plot the Pareto frontiers of GLUE scores against hardware performance measures 
for each platform in Fig.~\ref{fig:pareto}. We obtain these frontiers by optimizing each hardware performance measure 
for every device using its surrogate model obtained using ProTran (with GLUE scores predicted by the surrogate model
obtained using FlexiBERT 2.0). Fig.~\ref{fig:pareto}(a) shows the Pareto frontiers for GLUE scores plotted against 
inference latency. A100, due to its high batch size of 128 (see Table~\ref{tab:sample_sizes}), has much lower latency than embedded platforms. The 
Apple M1 GPU has around 2$\times$ higher average latency (along the frontier). Other devices have much higher 
latencies, running into hundreds of seconds averaged per sequence. Hence, if latency is a hard constraint, 
e.g., in a real-time language translation application, a designer can switch to other devices if a particular device 
does not meet the constraint. These profiled curves thus obviate the need for HW-NAS on every platform in future 
edge deployments.
Fig.~\ref{fig:pareto}(b) plots the Pareto frontiers for GLUE scores against energy consumption. The Apple M1 GPU 
has the lowest energy consumption, slightly lower than that of the A100 GPU. Fig.~\ref{fig:pareto}(c) shows the 
Pareto frontiers of the GLUE scores plotted against peak power draw. Intel NCS has the lowest power consumption. 
The nearly-vertical frontier indicates similar power consumption characteristics for diverse transformer models. 
On the other hand, the A100 GPU has drastically higher peak power curves, running into hundreds of Watts. A common 
trend in all three plots is that the GLUE score generally increases with increasing hardware performance measures. 
This correlates with results presented for deeper and wider architectures in Figs.~\ref{fig:contour_a100}(a) 
and \ref{fig:contour_a100}(b).

\subsection{Co-design of the Transformer and Edge Platform}

\begin{figure}
    \centering
    \includegraphics[width=\linewidth]{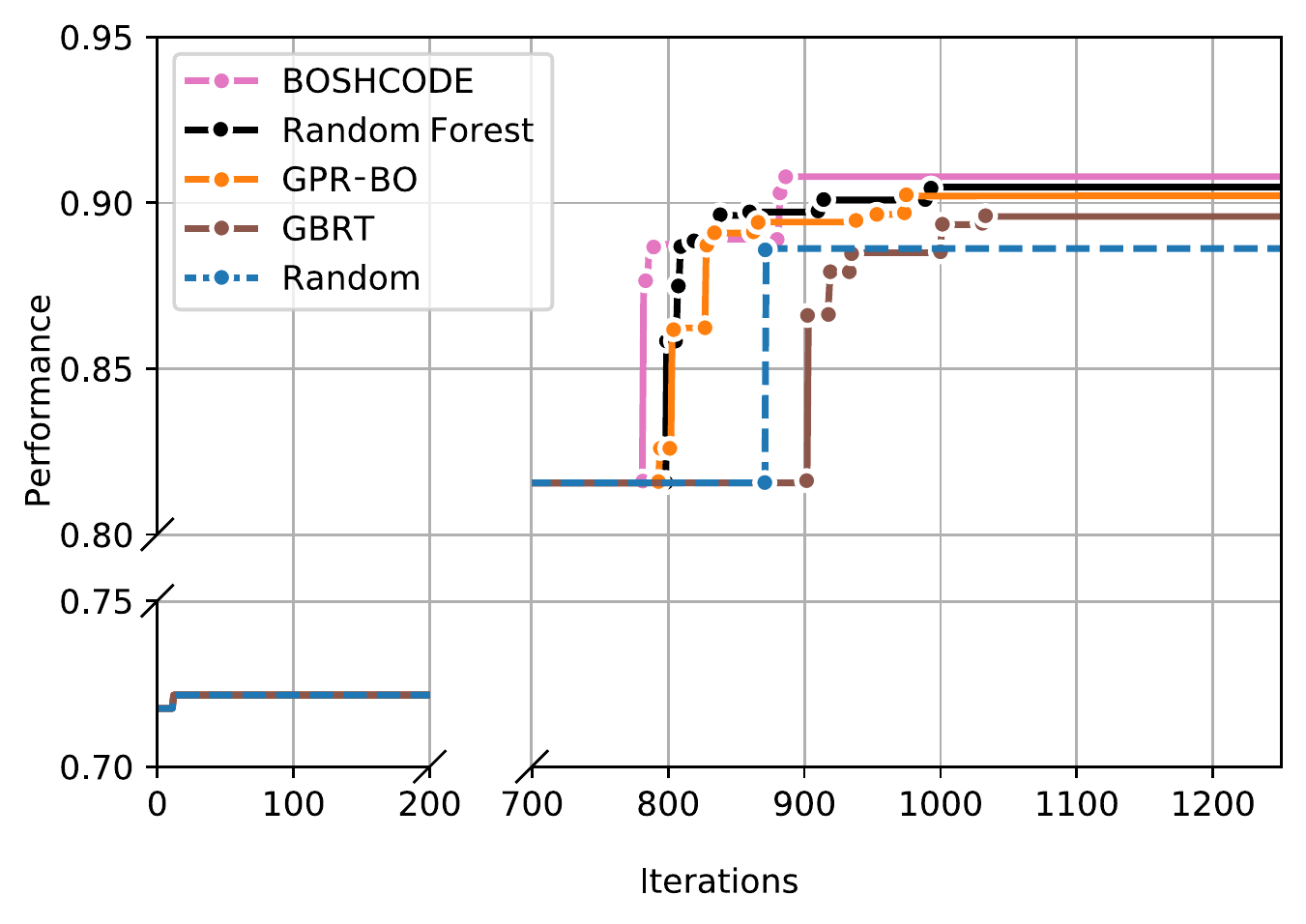}
    \caption{Convergence plots for co-design experiments using BOSHCODE and various baselines.}
    \label{fig:codesign_convergence}
\end{figure}

\begin{table}[]
\caption{Design choices of the converged ET model obtained using BOSHCODE.}
\centering
\begin{tabular}{@{}l@{\hskip 0.2in}l@{\hskip 0.2in}|@{\hskip 0.2in}l@{}}
\toprule
\textbf{Hyperparameter}                   &        & \textbf{Value}            \\ \midrule
\textbf{\multirow{8}{*}{Encoder Layer 1}} & $h^1$  & 256              \\ [1mm]
                                 & \#SA-SDP & 2                \\ [1mm]
                                 & \#SA-WMA & 2                \\ [1mm]
                                 & \#LT-DFT & 2                \\ [1mm]
                                 & \#LT-DCT & 1                \\ [1mm]
                                 & \#DSC-5  & 1                \\ [1mm]
                                 & \#DSC-9  & 4                \\ [1mm]
                                 & FF     & 4096, 3072, 4096 \\ \midrule
\textbf{\multirow{6}{*}{Encoder Layer 2}} & $h^2$  & 256              \\ [1mm]
                                 & \#SA-WMA & 2                \\ [1mm]
                                 & \#LT-DFT & 3                \\ [1mm]
                                 & \#DSC-5  & 2                \\ [1mm]
                                 & \#DSC-9  & 5                \\ [1mm]
                                 & FF     & 256, 256, 2048   \\ \bottomrule
\end{tabular}
\label{tab:trancode_design_choices}
\end{table}

\begin{table*}
\caption{Ablation analysis and baseline comparisons for our proposed EdgeTran + GPTran framework.}
\centering
\resizebox{\linewidth}{!}{
\begin{tabular}{@{}lccccccccc@{}}
\toprule
\textbf{Method}                  & \textbf{\begin{tabular}[c]{@{}c@{}}Hardware-\\ Aware\end{tabular}} & \textbf{\begin{tabular}[c]{@{}c@{}}Flex. \\ Layers\end{tabular}} & \textbf{\begin{tabular}[c]{@{}c@{}}Fine-grained \\ Search\end{tabular}} & \textbf{Platform}     & \textbf{\#Params.} & \textbf{\begin{tabular}[c]{@{}c@{}}GLUE Score \\ (\%)\end{tabular}} & \textbf{\begin{tabular}[c]{@{}c@{}}Latency \\ (ms/seq)\end{tabular}} & \textbf{\begin{tabular}[c]{@{}c@{}}Energy \\ (J/seq)\end{tabular}} & \textbf{\begin{tabular}[c]{@{}c@{}}Peak Power \\ (W)\end{tabular}} \\ \midrule
BERT-Base~\cite{bert} & \xmark & \xmark & \xmark & A100 & 110M & 79.6 & 10.57 & 0.61 & 199.86 \\ \midrule
\multicolumn{10}{c}{\textbf{Baseline Comparison}} \\ \midrule
AutoTinyBERT~\cite{autotinybert} & \xmark & \xmark & \xmark & Raspberry Pi & 60.7M & 78.3 & 10,427.73 & 20.69 & 5.02 \\ [1mm]
HAT~\cite{hat_mit} & \cmark & \xmark & \xmark & Raspberry Pi & 96.0M & 77.1 & 12,351.61 & 38.21 & 4.95 \\ [1mm] 
\textbf{EdgeTran (HW-NAS; Ours)} & \cmark & \cmark & \xmark & Raspberry Pi & 41.4M & 77.9 & 10,371.90 & 18.62 & \textbf{4.65} \\ \midrule
\multicolumn{10}{c}{\textbf{Ablation Analysis}} \\ \midrule
\textcolor{black}{\textbf{EdgeTran (vanilla-NAS; Ours)}} & \textcolor{black}{\xmark} & \textcolor{black}{\cmark} &
\textcolor{black}{\xmark} & \textcolor{black}{M1 GPU} & \textcolor{black}{139.0M} & \textcolor{black}{\textbf{81.8}} & \textcolor{black}{31.45} & \textcolor{black}{0.583} & \textcolor{black}{23.13} \\ [1mm]
\textbf{EdgeTran (w/o al. unc.; Ours)} & \cmark & \cmark & \xmark & M1 GPU & 43.8M & 77.6 & 14.56 & 0.118 & 21.44 \\ [1mm]
\textbf{EdgeTran (ET; Ours)} & \cmark & \cmark & \xmark & M1 GPU & 42.1M & 79.2 & 9.21 & 0.063 & 18.83 \\ [1mm]
\textbf{EdgeTran + GPTran (ET$^*$; Ours)} & \cmark & \cmark & \cmark & M1 GPU & \textbf{39.6M} & 80.4 & \textbf{8.98} & \textbf{0.061} & 18.47 \\ \bottomrule
\end{tabular}}
\label{tab:baseline_comp}
\end{table*}

Now that we have characterized the performance of each hardware platform, we experiment with co-design of the 
transformer model and edge-AI device. This entails running our proposed BOSHCODE pipeline along with several 
black-box optimization methods. Fig.~\ref{fig:codesign_convergence} shows how performance (see Eq.~\ref{eqn:perf_metric}) converges for different 
co-design schemes. These schemes include random search, gradient-boosted regression trees (GBRT), GPR-BO
that approximates performance through Gaussian process regression and optimizes it through the 
L-BFGS method~\cite{lbfgs}, and random forest that fits various randomized decision trees over sub-samples of the dataset. We employ a UCB estimate for optimization in all methods 
(except Random). As can be seen from Fig.~\ref{fig:codesign_convergence}, BOSHCODE achieves the highest
performance. It yields the optimal transformer-device pair, namely ET-Apple M1 GPU.

Table~\ref{tab:trancode_design_choices} shows the architectural design choices made in ET. To optimize 
latency, ET uses only two encoder layers. However, to avoid losing performance, ET uses 12 attention 
heads in each encoder layer. Thus, BOSHCODE searches for a shallow but wide model to improve throughput while not 
incurring a performance penalty. The converged architecture is also highly heterogeneous, with diverse attention 
types in each layer, leveraging the modeling capabilities of each operation type.

\begin{figure}
    \centering
    \includegraphics[width=\linewidth]{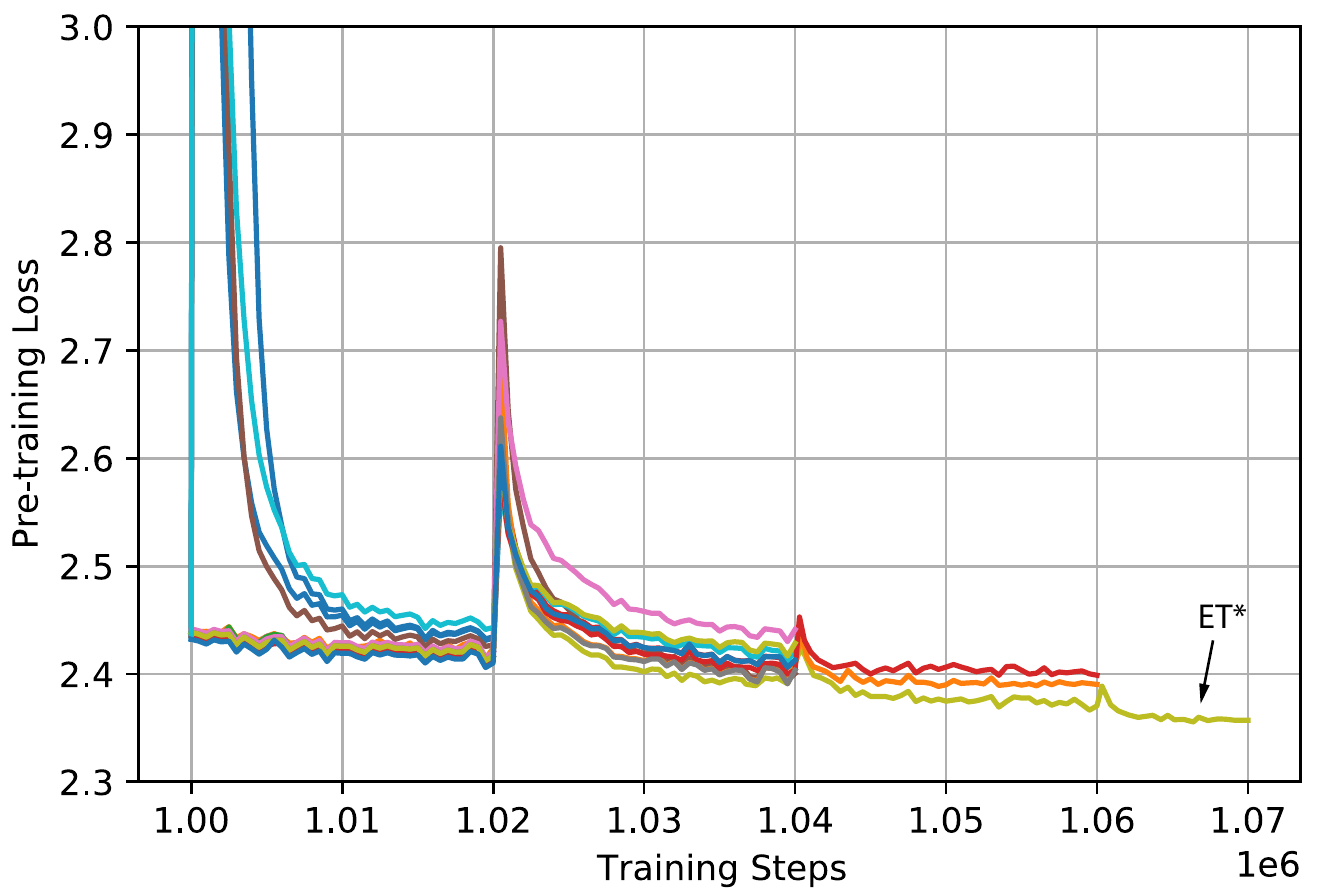}
    \caption{Loss curves for various children of ET while employing block-level growth and pruning.}
    \label{fig:gptran_loss}
\end{figure}

\subsection{GPTran Loss Curves}

GPTran optimizes the pre-training masked language modeling (MLM) loss while employing block-level growth and pruning 
on the given transformer model. Fig.~\ref{fig:gptran_loss} presents the MLM loss~\cite{bert} under different modes in the 
GPTran pipeline. ET$^*$, the model obtained after running GPTran on ET, has the lowest loss, 1.4\% lower than that 
of ET, while requiring 5.9\% fewer parameters than ET. Note how GPTran backtracks twice, after pruning two models 
(red and orange colors), while finally converging to ET$^*$ (olive color).

\subsection{Ablation Analysis and Baseline Comparison}

We now present an ablation analysis and baseline comparisons for our proposed framework. Table~\ref{tab:baseline_comp} 
compares EdgeTran and GPTran with the baselines mentioned in Section~\ref{sec:baselines}. The HW-NAS method employed in 
EdgeTran achieves a 2.4$\times$ smaller model, with 0.8\% higher GLUE score, 11.2\% lower latency, 25.1\% lower energy 
consumption, and 6.1\% lower peak power draw on the same platform, relative to HAT. 
The table then presents the results from the ablation experiments. \textcolor{black}{First, we test vanilla-NAS, where we block all gradients to the edge device and only search for a transformer model that maximizes the GLUE score. We evaluate the obtained model on the M1 GPU. The obtained model attains the highest GLUE score of 81.8, which is only 0.4 less than that of BERT-Large. However, this model is much smaller (139M against 345M parameters of BERT-Large).} Then, we test 
EdgeTran with co-optimization of the model and the edge device, however, without
leveraging aleatoric uncertainty (i.e., without the NPN model in BOSHCODE). We then show the performance of the
model-device pair from simultaneous co-design by EdgeTran (ET on M1 GPU). This leverages BOSHCODE, the surrogate models for the FlexiBERT
2.0 design space, and those from ProTran for every edge device. Finally, we present the results after running block-level
growth and pruning on the output transformer model obtained using BOSHCODE in EdgeTran (resulting in ET$^*$). GPTran reduces the model size and 
improves its performance on the GLUE benchmark. Reducing the model size also helps improve the hardware performance measures. \textcolor{black}{As seen from the table, hardware-software co-design results in substantial gains in accuracy, energy efficiency, and peak power consumption, compared to hand-designed models (e.g., BERT-Base) or vanilla NAS (without considering factors related to hardware).}
The resultant model-device pair yields a 2.8$\times$ smaller model with 0.8\% higher GLUE score, 15.0\% lower latency, 
10.0$\times$ lower energy, and 10.8$\times$ lower peak power than the baseline model (BERT-Base) on the A100 GPU.

\section{\textcolor{black}{Discussion}}
\label{sec:discussion}

\textcolor{black}{In this section we discuss the implications of the proposed work along with future work directions.}

\subsection{\textcolor{black}{Model Scheduling}}

\textcolor{black}{The latency and hardware utilization, which also affect net energy consumption and peak power draw, depend on how one schedules the ML model. Different scheduling strategies can leverage data reuse in distinct ways. In this work, we leverage the scheduling strategy of the supported deep learning framework that evaluates a given transformer model on the target hardware platform. This limits the scope of optimizing data reuse and resultant gains in latency and 
energy consumption. Extending the application of the FlexiBERT 2.0 design space to custom-designed 
accelerators~\cite{acceltran} could provide the flexibility of choosing among different scheduling decisions and 
dataflows. We leave these extensions to future work.}

\subsection{\textcolor{black}{Memory Utilization}}

\textcolor{black}{We fix the batch size when evaluating models (based on the largest transformer in the design space). 
Smaller models could see much lower memory utilization with a fixed batch size. However, adding flexibility for 
dynamic batch sizes would change latency, energy, and peak power draw profiles. This would require separate surrogate 
models for each set of batch sizes while evaluating a given hardware platform. Memory usage optimization can be 
incorporated into the ProTran framework, and we leave this extension to future work.}

\subsection{\textcolor{black}{Dynamic Workloads}}

\textcolor{black}{In this work, we assume that the target hardware platforms are dedicated to model inference alone. 
However, mobile edge devices are responsible for multiple functions and their workloads change dynamically. In such 
scenarios, the surrogate model can be dynamically updated based on the latest workload characteristics (such effects 
can also be tested {\em a priori}). One could use such heuristics and dynamic surrogate models in these settings.}

\textcolor{black}{EdgeTran only implements a static model on the edge device. Taking inspiration from recent works 
that run dynamic inference~\cite{dynamic_hat} on the given edge platform, we could perform inference with only a part 
of the transformer at runtime. We could also dynamically scale the model itself based on available resources.}

\subsection{\textcolor{black}{Pre-profiled Surrogate Models}}

\textcolor{black}{FlexiBERT 2.0 took 100 GPU-days to obtain the surrogate model. ProTran can profile a hardware platform 
within a few hours. BOSHCODE, on the other hand, uses the surrogate models (FlexiBERT 2.0 and ProTran, as shown in 
Fig.~\ref{fig:edgetran_framework}). ProTran took the maximum time to profile Raspberry Pi (7 hours). Owing to the 
lightweight surrogate models, running the co-design pipeline is very fast. With these pre-profiled surrogate models, 
implementing co-design takes a few minutes. One only needs to profile a new hardware platform with ProTran and then 
use the FlexiBERT 2.0 surrogate model to implement co-design in minutes. Thus, the FlexiBERT 2.0 surrogate model is 
very powerful in searching for new model-device pairs for a given set of design parameters 
($\alpha$, $\beta$, $\gamma$, and $\epsilon$). This highlights the advantages of the proposed strategy in terms
of search efficiency for future designs.}

\section{Conclusion}
\label{sec:conclusion}

This work presented multiple frameworks for efficient co-design of the transformer architecture and the edge-AI platform. 
First, we proposed ProTran, a framework to profile various hardware performance measures, including latency, energy, and peak 
power draw for a diverse set of transformer architectures on an inclusive set of mobile platforms for edge-AI. We also 
presented an expanded design space of transformer architectures, FlexiBERT 2.0, for a thorough evaluation of diverse mobile-friendly models. We leveraged BOSHCODE~\cite{codebench}, which employs training of surrogate models through active learning for efficient co-design of the 
transformer model and edge device. Finally, we optimized the output model with a hardware-aware post-processing step, GPTran, 
that employs fast block-level grow-and-prune. Based on the trained surrogate models, the optimal architectures while
running on mobile platforms achieve up to 22.0\% lower latency, 3.7$\times$ lower energy, and 48.8$\times$ lower peak power 
draw when compared with an off-the-shelf server-side GPU. Our method achieves a 2.8$\times$ smaller model with 0.8\% higher 
GLUE score, 15.0\% lower latency, 10.0$\times$ lower energy, and 10.8$\times$ lower peak power than the baseline model run on a 
server-side GPU.

\section*{Acknowledgments}

We performed the simulations presented in this article on computational resources managed and supported by Princeton Research 
Computing at Princeton University. We obtained the edge platforms with support from the Department of Electrical and Computer 
Engineering at Princeton University.

\bibliographystyle{IEEEtran}
{\footnotesize
\bibliography{IEEEabrv, biblio}}

\begin{IEEEbiography}[{\includegraphics[width=1in,height=1.5in,clip,keepaspectratio]{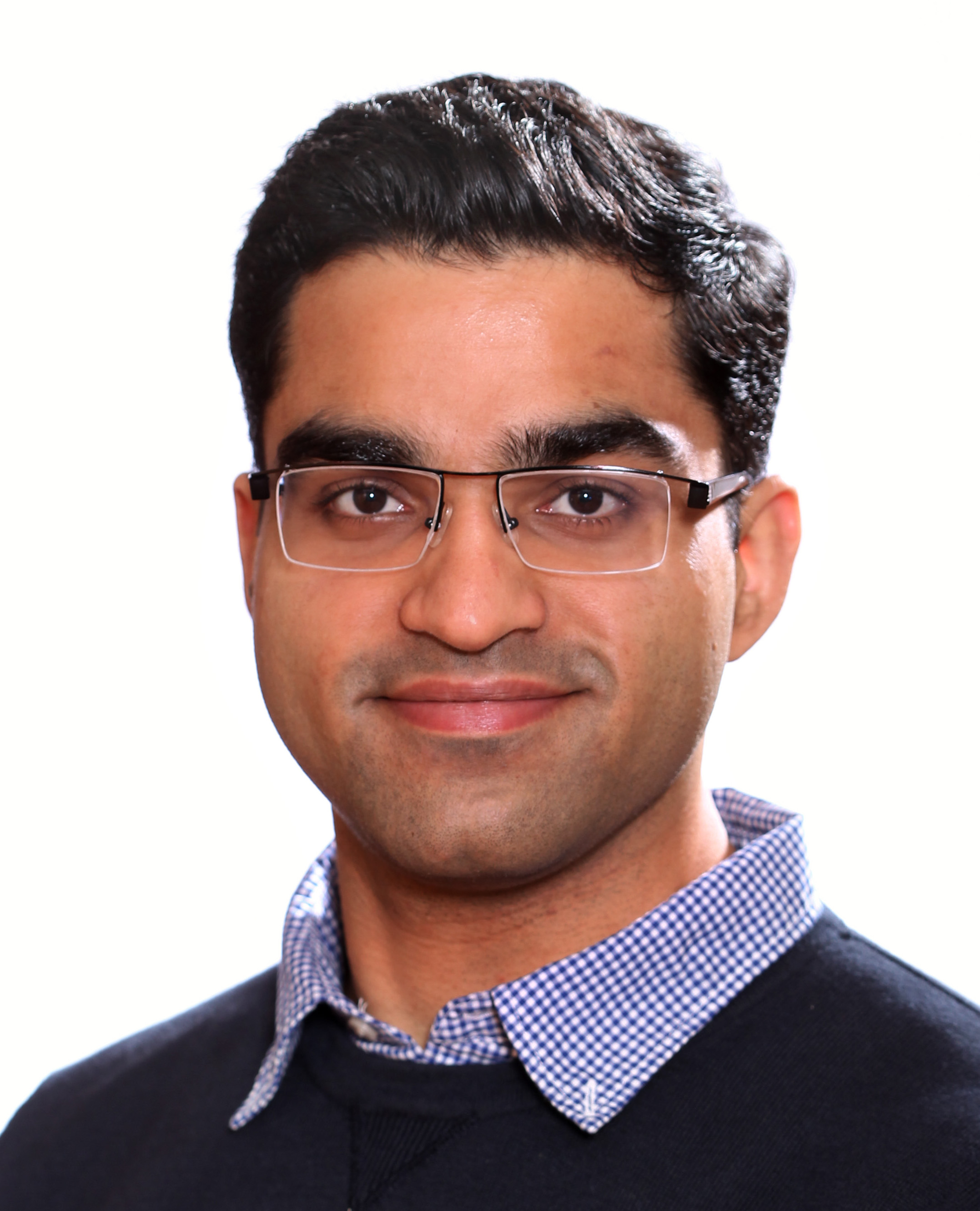}}]{Shikhar Tuli}
received the B. Tech. degree in electrical and electronics engineering from the Indian Institute of Technology (IIT) 
Delhi, India, with a department specialization in very large-scale integration (VLSI) and embedded systems. He is 
currently pursuing a Ph.D. degree at Princeton University in the department of electrical and computer engineering. 
His research interests include deep learning, edge artificial intelligence (AI), hardware-software co-design, 
brain-inspired computing, and smart healthcare.
\end{IEEEbiography}

\begin{IEEEbiography}[{\includegraphics[width=1in,height=1.5in,clip,keepaspectratio]{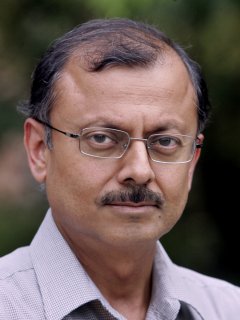}}]{Niraj K. Jha}
(Fellow, IEEE) received the B.Tech. degree in electronics and electrical communication engineering from 
IIT, Kharagpur, India, in 1981, and the Ph.D. degree in electrical engineering from the 
University of Illinois at Urbana–Champaign, Champaign, IL, USA, in 1985. 
He is a professor of electrical and computer engineering, Princeton University. 
He has co-authored five widely used books. He has published more than 470 papers (h-index: 83). 
He has received the Princeton Graduate Mentoring Award. His research has won 15 best paper awards, six award 
nominations, and 25 patents. He was given the Distinguished Alumnus Award by IIT, Kharagpur, in 2014. He has served 
as the Editor-in-Chief of TVLSI and an associate editor of several IEEE Transactions and other journals. He has 
given several keynote speeches in the areas of nanoelectronic design/test, smart healthcare, and cybersecurity.  
He is a fellow of ACM. His research interests include machine learning algorithms/architectures and smart healthcare. 
\end{IEEEbiography}

\end{document}